\documentclass[10pt,twocolumn,letterpaper]{article}

\usepackage[pagenumbers]{cvpr}
\usepackage{graphicx}
\usepackage{amsmath}
\usepackage{amssymb}
\newtheorem{assumption}{Assumption}
\usepackage{booktabs}
\usepackage{tabularx}
\usepackage[accsupp]{axessibility}
\usepackage[pagebackref,breaklinks,colorlinks]{hyperref}
\newcolumntype{C}{>{\centering\arraybackslash}X}
\makeatletter
\newcommand\footnoteref[1]{\protected@xdef\@thefnmark{\ref{#1}}\@footnotemark}
\makeatother
\def\cvprPaperID{10588}
\def\confName{CVPR}
\def\confYear{2022}

\begin{document}

\title{\vspace{-3mm}AR-NeRF: Unsupervised Learning of Depth and Defocus Effects
  \\from Natural Images with Aperture Rendering Neural Radiance Fields\vspace{-3mm}}

\author{Takuhiro Kaneko
  \vspace{2mm}\\
  NTT Communication Science Laboratories, NTT Corporation}

\twocolumn[{
  \renewcommand\twocolumn[1][]{#1}
  \maketitle
  \vspace{-7mm}
  \begin{center}
    \includegraphics[width=\textwidth]{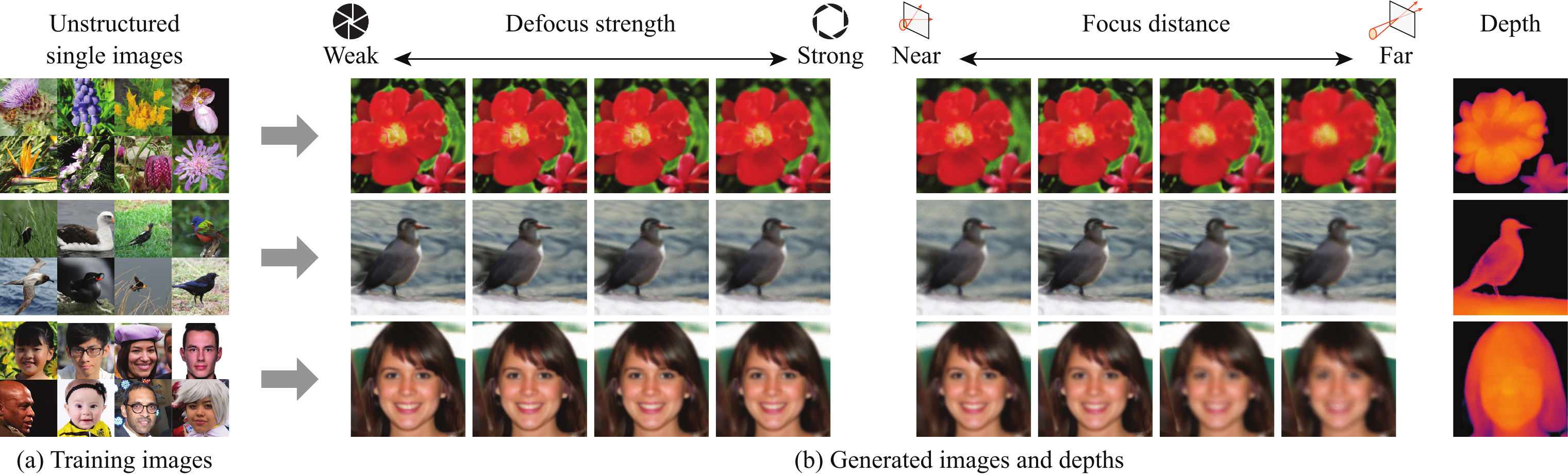}
  \end{center}
  \vspace{-5mm}
  \captionof{figure}{\textbf{Unsupervised learning of depth and defocus effects from unstructured (and view-limited) natural images.}
    (a) During training, we used \textit{only} a collection of unstructured (and view-limited) single natural images and did \textit{not} use any supervision (e.g., ground-truth depth, pairs of multiview images, defocus supervision, or pretrained models).
    (b) After training, our model, called \textit{AR-NeRF}, can generate sets of \textit{images} and \textit{depths}.
    In particular, in the generation of an image, AR-NeRF can adjust the \textit{defocus strength} and \textit{focus distance} intuitively and continuously using photometric constraints.
    The project page is available at \url{https://www.kecl.ntt.co.jp/people/kaneko.takuhiro/projects/ar-nerf/}.}
  \label{fig:concept}
  \vspace{5mm}
}]

\begin{abstract}
  Fully unsupervised 3D representation learning has gained attention owing to its advantages in data collection. A successful approach involves a viewpoint-aware approach that learns an image distribution based on generative models (e.g., generative adversarial networks (GANs)) while generating various view images based on 3D-aware models (e.g., neural radiance fields (NeRFs)). However, they require images with various views for training, and consequently, their application to datasets with few or limited viewpoints remains a challenge. As a complementary approach, an aperture rendering GAN (AR-GAN) that employs a defocus cue was proposed. However, an AR-GAN is a CNN-based model and represents a defocus independently from a viewpoint change despite its high correlation, which is one of the reasons for its performance. As an alternative to an AR-GAN, we propose an aperture rendering NeRF (AR-NeRF), which can utilize viewpoint and defocus cues in a unified manner by representing both factors in a common ray-tracing framework. Moreover, to learn defocus-aware and defocus-independent representations in a disentangled manner, we propose aperture randomized training, for which we learn to generate images while randomizing the aperture size and latent codes independently. During our experiments, we applied AR-NeRF to various natural image datasets, including flower, bird, and face images, the results of which demonstrate the utility of AR-NeRF for unsupervised learning of the depth and defocus effects.
\end{abstract}

\vspace{-2mm}
\section{Introduction}
\label{sec:introduction}

Natural images are 2D projections of the 3D world.
Solving the inverse problem, i.e., understanding the 3D world from natural images, is a principal challenge in computer vision and graphics and has been actively studied in various fields owing to its diverse applications, such as environmental understanding in robotics, content creation in advertisements, and photo editing in the arts.

After collecting pairs of 2D and 3D data or sets of multiview images, a successful approach is to learn the 3D predictor using direct or photometric-driven supervision.
This approach demonstrates promising results in terms of fidelity. However, the collection of such data is often difficult or impractical.
To reduce the collection costs, learning from single images (i.e., from a dataset that includes a single image per training instance) has been actively studied.

To obtain clues under such setting, several studies~\cite{AKanazawaCVPR2018,MWangIJCV2019,BGecerCVPR2019,SSanyalCVPR2019,JShangECCV2020} have introduced object-specific shape models, including 3DMM~\cite{VBlanzSIGGRAPH1999} and SMPL~\cite{MLoperTOG2015}, and searched for solutions within the shape model constraints.
Other studies have utilized auxiliary information such as 2D keypoints~\cite{LTranCVPR2018,AKanazawaECCV2018} or 2D silhouettes~\cite{PHenzlerICCV2019,WChenNeurIPS2019,XLiECCV2020,SGoelECCV2020} to simplify the problem by aligning the object parts or separating the target objects from the background.
These studies also demonstrate remarkable results; however, the construction of the shape model is not always easy and narrows the applicable objects, and auxiliary information incurs extra costs in terms of data collection.

To alleviate such restrictions, a \textit{fully} unsupervised approach, which learns 3D representations from single images without \textit{any} additional supervision (including auxiliary information and pre-trained models), has gained attention.
Under this setting, the viewpoint is a principal clue, which typical methods utilize by learning an image distribution using a generative model (e.g., a generative adversarial network (GAN)~\cite{IGoodfellowNIPS2014}) while generating various viewpoint images based on viewpoint-aware 3D models, such as voxels~\cite{PHenzlerICCV2019,TNguyenICCV2019,TNguyenNeurIPS2020}, primitives~\cite{YLiaoCVPR2020}, and neural radiance fields (NeRFs)~\cite{BMildenhallECCV2020,KSchwarzNeurIPS2020,EChanCVPR2021,MNiemeyerCVPR2021,MNiemeyer3DV2021,JGuICLR2022}.
This allows learning a viewpoint-aware 3D representation; however, owing to the diverse viewpoints needed, the application to a dataset in which viewpoint cues are limited or unavailable without the use of a preprocessing (e.g., natural flower or bird images, as shown in Figure~\ref{fig:concept}) remains a challenge.

As a complement to a viewpoint cue, an aperture rendering GAN (AR-GAN)~\cite{TKanekoCVPR2021b} was proposed to exploit a defocus cue by equipping the aperture rendering~\cite{PPSrinivasanCVPR2018} on top of a CNN GAN.
This constraint allows the learning of both depth and depth-of-field (DoF) effects in an unsupervised manner.
However, as a limitation, an AR-GAN employs the defocus cue independently from the viewpoint cue and cannot utilize both factors jointly despite these two factors being highly correlated with the ability to help each other.\footnote{More precisely, in ~\cite{TKanekoCVPR2021b}, the combinations of an AR-GAN and viewpoint-aware GANs (particularly, HoloGAN~\cite{TNguyenICCV2019} and RGBD-GAN~\cite{ANoguchiICLR2020}) are provided.
  These models can learn the defocus and viewpoint-aware representations \textit{simultaneously} but \textit{individually}; therefore, such models cannot utilize the learning of one representation for the learning of another.}
Consequently, the quality of the depth prediction when using AR-GAN remains limited.

We thus aim to construct a \textit{unified model} that can leverage defocus and viewpoint cues jointly by considering the application of unsupervised learning of the 3D representation (particularly \textit{depth} and \textit{defocus effects}) from natural unstructured (and view-limited) images (Figure~\ref{fig:concept}).
To achieve this, we propose a new extension of NeRF called \textit{aperture rendering NeRF (AR-NeRF)}, which can represent defocus effects and viewpoint changes in a unified manner by representing both factors through a common ray-tracing framework.
More precisely, in contrast to the standard NeRF, which represents each pixel using a single ray under the \textit{pinhole camera} assumption, AR-NeRF employs an \textit{aperture camera}~\cite{Shirley2020RTW1} that represents each pixel using a collection of rays that converge at the focus plane and whose scale is determined according to the aperture size.
Through such modeling, we can represent both viewpoint changes and defocus effects by simply changing the inputs and the integration of the implicit function (multilayer perceptron (MLP)), which converts the point position and view direction into the RGB color and volume density.
Consequently, through training, we can optimize the MLP while reflecting both factors.

Moreover, to disentangle defocus-aware and defocus-independent representations in an unsupervised manner, we introduce \textit{aperture randomized training}, in which we learn to generate images in a GAN framework while changing the aperture size and latent codes both randomly and independently.
A similar technique is commonly used in viewpoint-aware representation learning~\cite{PHenzlerICCV2019,TNguyenICCV2019,TNguyenNeurIPS2020,YLiaoCVPR2020,BMildenhallECCV2020,KSchwarzNeurIPS2020,EChanCVPR2021,MNiemeyerCVPR2021,MNiemeyer3DV2021,JGuICLR2022}, and this training is useful for disentangling the effect of the corresponding factor from latent codes.

We applied AR-NeRF to natural image datasets, including view-limited (Oxford Flowers~\cite{MENilsbackICVGIP2008} (flower) and CUB-200-2011~\cite{CWahCUB2002011} (bird)) datasets and a dataset with various views (FFHQ~\cite{TKarrasCVPR2019} (face)), and demonstrated that AR-NeRF is better than or comparable to the baseline models, including a state-of-art fully unsupervised depth-learning model (i.e., AR-GAN~\cite{TKanekoCVPR2021b}) and generative NeRF (particularly pi-GAN~\cite{EChanCVPR2021}), in terms of the \textit{depth} prediction accuracy.
We also demonstrated that AR-NeRF can manipulate the \textit{defocus effects} (i.e., defocus strength and focus distance) intuitively and continuously while retaining the image quality, whereas AR-GAN has difficulty doing so.

Overall, our contributions can be summarized as follows:
\begin{itemize}
  \vspace{-0.5mm}
  \setlength{\parskip}{2pt}
  \setlength{\itemsep}{2pt}
\item To achieve an unsupervised learning of the depth and defocus effects, we propose a new extension of NeRF called \textit{AR-NeRF}, which can employ viewpoint and defocus cues in a unified manner by representing both factors in a common ray-tracing framework.
\item To disentangle defocus-aware and defocus-independent representations under unsupervised conditions, we introduce \textit{aperture randomized training}, by which we learn to generate images while changing the aperture size and latent codes both randomly and independently.
\item We empirically demonstrate the utility of AR-NeRF for the unsupervised learning of the \textit{depth} and \textit{defocus effects} using various natural image datasets, including view-limited (flower and bird) datasets and a dataset of various views (face).
  We provide detailed analyses and extended results in the Appendices.\footnote{The project page is available at \url{https://www.kecl.ntt.co.jp/people/kaneko.takuhiro/projects/ar-nerf/}.}
\end{itemize}

\section{Related work}
\label{sec:related_work}

\noindent\textbf{Implicit neural representations.}
Owing to their 3D-aware, continuous, and memory-efficient nature, implicit neural representations have gained attention in both learning-based 3D ~\cite{KGenovaICCV2019,JParkCVPR2019,MAtzmonCVPR2020,MMichalkiewiczICCV2019,LMeschederCVPR2019,ZChenCVPR2019,ZChenICCV2019,MNiemeyerICCV2019,MOechsleICCV2019,SSaitoICCV2019} and scene ~\cite{CJiangCVPR2020,RChabraECCV2020,JChibaneNeurIPS2020,SPengECCV2020} reconstructions. 
Typical representations are supervised using 3D data; however, to eliminate the need for 3D supervision, the incorporation of differentiable rendering has also been proposed~\cite{SLiuNeurIPS2019,SLiuCVPR2020,MNiemeyerCVPR2020,VSitzmannNeurIPS2019,LYarivNeurIPS2020}.
The most relevant model is NeRF~\cite{BMildenhallECCV2020}, which combines implicit neural representations with volume rendering for a novel view synthesis.
Our AR-NeRF is based on NeRF and obtains a \textit{viewpoint-aware} functionality by inheriting it. However, to obtain the \textit{defocus-aware} functionality, AR-NeRF employs an \textit{aperture camera} model instead of a \textit{pinhole camera} model, which is typically used in NeRF.
Moreover, the described studies aimed to learn a single network per scene using a \textit{set of multiview} images, whereas we aimed to construct a generative model from the collection of \textit{unstructured single} images.
Owing to this difference, we do not aim to compare AR-NeRF with NeRF in this study; however, reimporting our idea (i.e., the usage of an aperture camera) to the original task remains for future research.

\smallskip\noindent\textbf{Generative adversarial networks.}
GANs~\cite{IGoodfellowNIPS2014} have shown remarkable results in 2D image modeling through a series of advancements (e.g.,~\cite{ABrockICLR2019,TKarrasICLR2017,TKarrasCVPR2019,TKarrasCVPR2020,TKarrasNeurIPS2021}).
A strong property of GANs is their ability to learn the data distribution through random sampling without directly defining the distribution.
This property allows GANs to learn the data distribution through measurements~\cite{ABoraICLR2018,APajotICLR2018,SLiICLR2019,TKanekoCVPR2020,TKanekoCVPR2021a} and architectural constraints~\cite{XWangECCV2016,JWuNIPS2016,CVondrickNIPS2016,JYangICLR2017,JYZhuNeurIPS2018}.
Using the same logic, unsupervised 3D-aware GANs~\cite{TNguyenICCV2019,PHenzlerICCV2019,ASzaboArXiv2019,ANoguchiICLR2020,YLiaoCVPR2020,BMildenhallECCV2020,KSchwarzNeurIPS2020,TNguyenNeurIPS2020,EChanCVPR2021,MNiemeyerCVPR2021,MNiemeyer3DV2021,JGuICLR2022,TKanekoCVPR2021b} have succeeded in learning 3D-aware representations by incorporating 3D-2D projection modules and/or 3D-aware constraints.
More specifically, most studies address \textit{viewpoint-aware} representation learning using 3D representations based on voxels~\cite{PHenzlerICCV2019,TNguyenICCV2019,TNguyenNeurIPS2020}, primitives~\cite{YLiaoCVPR2020}, and NeRFs~\cite{BMildenhallECCV2020,KSchwarzNeurIPS2020,EChanCVPR2021,MNiemeyerCVPR2021,MNiemeyer3DV2021,JGuICLR2022}, and a few studies~\cite{TKanekoCVPR2021b} have addressed the learning of \textit{defocus-aware} representations.
Herein, we introduce a unified model that can jointly leverage both \textit{defocus} and \textit{viewpoint} cues to strengthen the latter category model.
We demonstrate the utility of their joint usage in the experiments described in Section~\ref{subsec:ablation_study}.

\smallskip\noindent\textbf{Learning 3D representations from single images.}
As discussed in Section~\ref{sec:introduction}, to eliminate the cost of collecting 3D data or multiview images, the learning of a 3D representation from single images has garnered attention.
Promising approaches involve the use of shape models~\cite{AKanazawaCVPR2018,MWangIJCV2019,BGecerCVPR2019,SSanyalCVPR2019,JShangECCV2020} and the incorporation of auxiliary information such as 2D keypoints~\cite{LTranCVPR2018,AKanazawaECCV2018} or 2D silhouettes~\cite{PHenzlerICCV2019,WChenNeurIPS2019,XLiECCV2020,SGoelECCV2020}.
Although such approaches have yielded remarkable results, the requirement for shape models or auxiliary information remains a bottleneck.
To eliminate this bottleneck, a \textit{fully} unsupervised learning approach based on generative models has been actively studied.
The learning targets differ according to the studies applied, and to date, the unsupervised learning of \textit{viewpoints}~\cite{TNguyenICCV2019,PHenzlerICCV2019,ASzaboArXiv2019,ANoguchiICLR2020,YLiaoCVPR2020,BMildenhallECCV2020,KSchwarzNeurIPS2020,TNguyenNeurIPS2020,EChanCVPR2021,MNiemeyerCVPR2021,MNiemeyer3DV2021,JGuICLR2022,SWuCVPR2020,MSahasrabudheICCVW2019}, \textit{albedo}~\cite{SWuCVPR2020}, \textit{texture}~\cite{ASzaboArXiv2019}, \textit{light}~\cite{SWuCVPR2020}, \textit{3D meshes}~\cite{ASzaboArXiv2019,MSahasrabudheICCVW2019}, \textit{depth}~\cite{ANoguchiICLR2020,SWuCVPR2020,TKanekoCVPR2021b}, and \textit{defocus effects}~\cite{TKanekoCVPR2021b} has been proposed.
Among these approaches, AR-NeRF shares the motivation with AR-GAN~\cite{TKanekoCVPR2021b} with the aim of learning the \textit{depth} and \textit{defocus effects}.
However, as the main difference, an AR-GAN represents an aperture renderer in \textit{discretized} CNNs and is specific to the learning of the \textit{depth} and \textit{defocus effects}, whereas our AR-NeRF represents an aperture renderer with \textit{continuous} radiance fields and can explain and utilize other ray-tracing-related phenomena (e.g., \textit{viewpoints}) in a unified manner.
We empirically demonstrate these merits in Section~\ref{subsec:comparative_study}.

\smallskip\noindent\textbf{Learning of depth and defocus effects.}
There is a large body of studies conducted on depth learning.
Representative approaches involve training the depth predictor using pairs or sets of data, such as image and depth pairs~\cite{DEigenNIPS2014,FLiuTPAMI2015,ILaina3DV2016,YKuznietsovCVPR2017,DXuCVPR2017,HFuCVPR2018}, multiview image pairs~\cite{RGargECCV2016,CGodardCVPR2017,KXianCVPR2020}, and consecutive frame sets~\cite{TZhouCVPR2017,ZYinCVPR2018,CWangCVPR2018}.
Defocus synthesis has also garnered interest in computer vision and graphics, and both model-based~\cite{JTBarronCVPR2015,DEJacobsSCTLTR2012,SWHasinoffICCV2007,XShenECCV2016,NWadhwaTOG2018} and learning-based~\cite{PPSrinivasanCVPR2018,LWangTOG2018,AIgnatovCVPRW2020,MQianECCVW2020} defocus synthesizers have been proposed.
Based on the high correlation between the depth and defocus strength, some studies~\cite{PPSrinivasanCVPR2018,SGurCVPR2019} have proposed learning the depth while reconstructing focused images from all-in-focus images under the assumption that pairs of focused and all-in-focus images are available for training.
Although our study is motivated by the success of such studies, the main difference is that we address a challenging but practically important situation in which there are \textit{no} training data available other than natural unstructured (and view-limited) images.
The latest model addressing this problem is an AR-GAN~\cite{TKanekoCVPR2021b}.
As stated previously, we investigate the quantitative and qualitative differences in Section~\ref{subsec:comparative_study}.

\section{Preliminaries}
\label{sec:preliminaries}

\subsection{GAN}
\label{subsec:gan}

We begin by describing the two previous work upon which our model is built.
The first is a GAN~\cite{IGoodfellowNIPS2014}, which learns the data distribution implicitly through a two-player min-max game using the following objective function:
\begin{flalign}
  \label{eqn:gan}
  \mathcal{L}_{\text{GAN}} = & \: \mathbb{E}_{\mathbf{I}^r \sim p^r(\mathbf{I})} [\log D(\mathbf{I}^r)]
  \nonumber \\
  + & \: \mathbb{E}_{\mathbf{z} \sim p^g(\mathbf{z})} [\log (1 - D(G(\mathbf{z})))],
\end{flalign}
where, given a latent code $\mathbf{z} \sim p^g(\mathbf{z})$, a generator $G$ generates an image $\mathbf{I}^g$ that fools the discriminator $D$ by minimizing $\mathcal{L}_{\text{GAN}}$, whereas $D$ distinguishes $\mathbf{I}^g$ from a real image $\mathbf{I}^r$ by maximizing $\mathcal{L}_{\text{GAN}}$.
Here, the superscripts $r$ and $g$ represent real and generated data, respectively.
Through adversarial training, $p^g(\mathbf{I})$ reaches close to $p^r(\mathbf{I})$.

\subsection{NeRF}
\label{subsec:nerf}

NeRF~\cite{BMildenhallECCV2020} (in particular, we consider generative variants~\cite{KSchwarzNeurIPS2020,EChanCVPR2021} relevant to our study) represents a scene using an MLP that takes the 3D position $\mathbf{x} \in \mathbb{R}^3$ and view direction $\mathbf{d} \in \mathbb{S}^2$ as inputs and predicts the RGB color $\mathbf{c}(\mathbf{x}, \mathbf{d}) \in \mathbb{R}^3$ and volume density $\sigma(\mathbf{x}) \in \mathbb{R}^+$.
More precisely, in~\cite{KSchwarzNeurIPS2020,EChanCVPR2021}, positional encoding~\cite{BMildenhallECCV2020,MTancikNeurIPS2020} and sine nonlinearity~\cite{VSitzmannNeurIPS2020} were used prior to or during the application of the MLP to encode positional information; however, we omit them for a general representation.
Moreover, in a generative variant, the MLP also takes the latent code $\mathbf{z} \in \mathbb{R}^{L_{\mathbf{z}}}$ as input to represent a variety of data.
However, this is omitted for simplicity.

NeRF employs a pinhole camera (Figure~\ref{fig:ray_tracing}(a)) and predicts the color of each pixel $\mathbf{C}(\mathbf{r})$ and the corresponding depth $Z(\mathbf{r})$ by integrating over a single camera ray $\mathbf{r}(t) = \mathbf{o} + t \mathbf{d}$ (where $\mathbf{o}$ and $\mathbf{d}$ are the camera origin and direction, respectively) within a distance $t \in [t_n, t_f]$ using the volume rendering equation~\cite{NMaxTVCG1995}:
\begin{flalign}
  \label{eqn:volume_rendering}
  \mathbf{C}(\mathbf{r}) & = \int_{t_n}^{t_f} T(t) \sigma(\mathbf{r}(t)) \mathbf{c}(\mathbf{r}(t), \mathbf{d}) dt,
  \nonumber \\
  Z(\mathbf{r}) & = \int_{t_n}^{t_f} T(t) \sigma(\mathbf{r}(t)) t dt,
  \nonumber \\
  \text{where } T(t) & = \exp \left(- \int_{t_n}^t \sigma(\mathbf{r}(s)) ds \right).
\end{flalign}
In practice, the integral is intractable; thus, a discretized form with stratified and hierarchical sampling~\cite{BMildenhallECCV2020} is used.

\section{Aperture rendering NeRF: AR-NeRF}
\label{sec:ar-nerf}

\subsection{Problem statement}
\label{subsec:problem_statement}

We first clarify the problem statement.
We address \textit{fully} unsupervised learning of the depth and defocus effects, where \textit{no} supervision or pretrained models are available and only a collection of unstructured single images are accessible during training.
Owing to the lack of explicit supervision, it is difficult to learn a \textit{conditional} model that can directly predict the depth and defocus effects from an input image.
As an alternative, we aim to construct an \textit{unconditional} generator $G(\mathbf{z})$ that can generate the image and depth as a set while varying the defocus effects.
Similar to viewpoint-aware representation learning, which requires a dataset that includes various view images to acquire the viewpoint cue, our defocus-aware representation learning requires a dataset that includes variously defocused images to obtain a defocus cue.
More formally, we impose the following assumption for the dataset:
\begin{assumption}
  \label{assumption:ar-nerf}
  The training images are captured using various aperture-sized cameras, and the dataset includes diversely defocused images.
\end{assumption}

Two factors affecting the defocus effects (as detailed in Section~\ref{subsec:aperture_rendering}) are the aperture size and focus distance (distance between the ray origin and the plane where all objects are in focus).
Hence, we can also impose the assumption of diversity of the focus distance instead of or in addition to Assumption~\ref{assumption:ar-nerf}.
However, under a practical scenario, the focused target tends to be fixed when the scene is determined.
Hence, in this study, only Assumption~\ref{assumption:ar-nerf} is imposed.\footnote{A similar assumption (diversity of the DoF settings) has also been introduced in an AR-GAN~\cite{TKanekoCVPR2021b}.
  However, it does not distinguish the effects of the two factors, and for a stricter assumption, we redefine it here.}

Note that we assume the existence of diversely defocused images but do \textit{not} assume the existence of their \textit{pairs/sets}.
We observed that this assumption is satisfied in a typical natural image dataset, as shown in Figure~\ref{fig:concept}.

\begin{figure}[t]
  \centering
  \includegraphics[width=0.99\columnwidth]{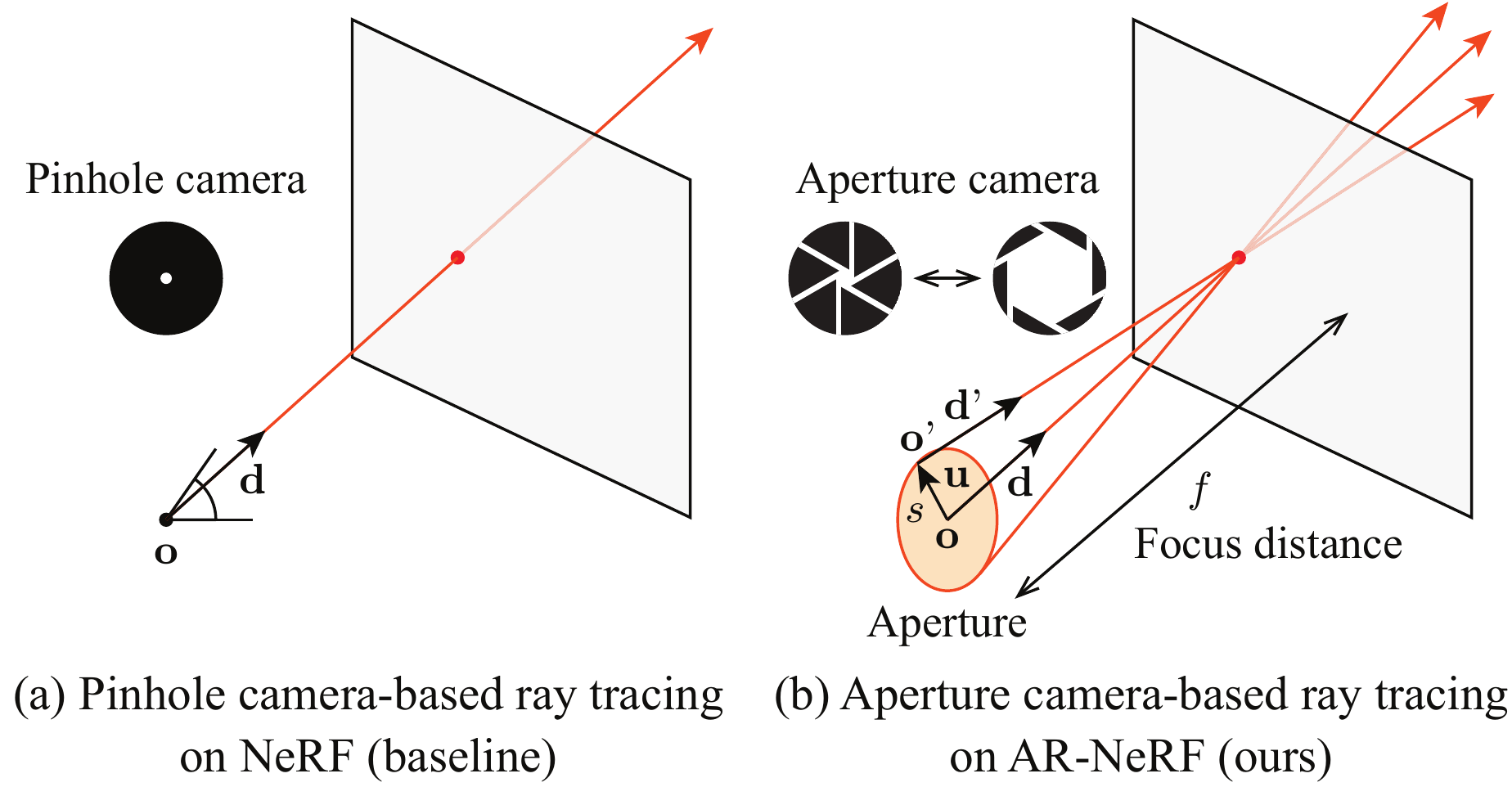}
  \vspace{-2mm}
  \caption{\textbf{Comparison of ray tracing on NeRF and AR-NeRF.}}
  \vspace{-4mm}
  \label{fig:ray_tracing}
\end{figure}

\subsection{Aperture rendering with NeRF}
\label{subsec:aperture_rendering}

As described in Section~\ref{subsec:nerf}, NeRF is a strongly 3D-aware model that can jointly represent an image and depth at the design level (Equation~\ref{eqn:volume_rendering}).
To utilize this strong property in our problem (Section~\ref{subsec:problem_statement}), we consider representing aperture rendering in a ray-tracing framework, which is the basis of NeRF.
This is achieved by replacing \textit{pinhole camera}-based ray tracing (Figure~\ref{fig:ray_tracing}(a)), which is used in a standard NeRF, with \textit{aperture camera}-based ray tracing~\cite{Shirley2020RTW1} (Figure~\ref{fig:ray_tracing}(b)).

For pinhole-camera-based ray tracing, we cast all rays from a single point $\mathbf{o}$.
By contrast, with aperture camera-based ray tracing, we cast rays from an aperture of radius $s$.
More formally, the origin of a ray from the aperture ($\mathbf{o}'$) is written as
\begin{flalign}
  \label{eqn:aperture_origin}
  \mathbf{o}' = \mathbf{o} + \mathbf{u},
\end{flalign}
where $|\mathbf{u}| \in [0, s]$, and the direction of $\mathbf{u}$ is orthogonal to $\mathbf{o}$.

A bundle of rays emitted from the aperture converges to a point on the plane at a focal distance of $f$.
Based on this definition, the direction of the ray from the aperture ($\mathbf{d}'$) is calculated as follows:
\begin{flalign}
  \label{eqn:aperture_direction}
  \mathbf{d}' = (\mathbf{o} + f \mathbf{d} - \mathbf{o}') / f,
\end{flalign}
Based on Equations~\ref{eqn:aperture_origin} and \ref{eqn:aperture_direction}, we can calculate the ray from the origin $\mathbf{o}'$, that is, $\mathbf{r}'(t) = \mathbf{o}' + t \mathbf{d}'$, and render the corresponding color $\mathbf{C}(\mathbf{r}')$ and depth $Z(\mathbf{r}')$ using volume rendering (Equation~\ref{eqn:volume_rendering}).
The final color and depth are calculated by integrating over $\mathbf{C}(\mathbf{r}')$ and $Z(\mathbf{r}')$ for all rays in $|\mathbf{u}| \in [0, s]$.\footnote{The depth does not need to be integrated, but we used this formulation to account for the ambiguity derived from the defocus blur.
  However, we empirically found that the effect on the depth accuracy is subtle.}
However, similar to volume rendering, the integral is intractable in practice; therefore, a discretized form is used.
More precisely, we generate a finite bundle of rays from the sampled $|\mathbf{u}| \in [0, s]$ and calculate the final output by taking the average of the corresponding $\mathbf{C}(\mathbf{r}')$ and $Z(\mathbf{r}')$.

\subsection{Aperture randomized training}
\label{subsec:aperture_randomized_training}

To learn defocus-aware and defocus-independent representations in a disentangled manner, we introduce \textit{aperture randomized training}, in which we learn to generate images by varying the aperture size and latent codes both randomly and independently.
More formally, we rewrite the GAN objective function (Equation~\ref{eqn:gan}) as follows:\footnote{The relevant training scheme is DoF-mixture learning~\cite{TKanekoCVPR2021b}, in which the image generation is learned while generating various defocused images.
  The main difference from our approach is that their method manipulates the depth scale instead of the aperture size because they cannot scale the aperture directly owing to the discretized formulation.
  Consequently, their learned depth is relative, and therefore they must carefully tune the value when changing the focus distance by adding an offset.}
\begin{flalign}
  \label{eqn:ar-nerf}
  \mathcal{L}_{\text{AR-NeRF}} = & \: \mathbb{E}_{\mathbf{I}^r \sim p^r(\mathbf{I})} [\log D(\mathbf{I}^r)]
  \nonumber \\
  + & \: \mathbb{E}_{\mathbf{z} \sim p^g(\mathbf{z}), s \sim p^g(s)} [\log (1 - D(G(\mathbf{z}, s)))],
\end{flalign}
where the latent code $\mathbf{z}$ and aperture size $s$ are sampled randomly and independently.
In practice, we represent $p^g(s)$ using a half-normal distribution and parameterize its standard deviation $\sigma_s$ to determine the range of aperture sizes in a data-driven manner.
As a side note, we represent the focus distance $f$, another variable in aperture rendering, using an MLP, taking $\mathbf{z}$ as the input under the assumption that $f$ is determined according to the rendered target.

As discussed in Section~\ref{subsec:aperture_rendering}, our aperture rendering has a strong 3D constraint based on ray tracing, and therefore when we train a model using Equation~\ref{eqn:ar-nerf}, $\mathbf{z}$ must capture the representations that are independent and robust to the change in defocus driven by the fluctuation of $s$.

\subsection{Advanced techniques for practice}
\label{subsec:advanced_techniques}

To the best of our knowledge, unsupervised learning of the depth and defocus effects is a relatively new task (e.g., the first attempt was at CVPR 2021~\cite{TKanekoCVPR2021b}), and practical techniques (in particular, those specific to NeRF) have yet to be sufficiently developed.
To advance this research direction, we discuss practical techniques considered for this task.

\smallskip\noindent\textbf{Representation of unbounded background with NeRF++.}
A typical generative NeRF~\cite{KSchwarzNeurIPS2020,EChanCVPR2021} renders an entire scene in a tightly bounded 3D space to efficiently model the foreground.
However, this strategy is problematic when training images with an unbounded background (e.g., bird images in Figure~\ref{fig:concept}).
In particular, this problem is critical in terms of learning the defocus effects because its strength is determined according to depth.
We cannot represent a strong defocus effect at the design level when using a tightly bounded 3D space.
To address this problem, we implemented a synthesis network using NeRF++~\cite{KZhangArXiv2020}, which is composed of a foreground NeRF in a unit sphere and background NeRF modeled using inverted sphere parameterization.
This implementation allows representing a strong defocus effect in a far background.
For a fair description, we note that concurrent approaches~\cite{MNiemeyer3DV2021,JGuICLR2022} have also incorporated NeRF++ for image generation to represent an unbounded background.

\smallskip\noindent\textbf{Learning depth and defocus effects with changes in viewpoint.}
Fully unsupervised learning of the depth and defocus effects is a challenging and ill-posed problem, although our aperture randomized training alleviates this difficulty.
To obtain a hint from another source, we jointly learn viewpoint-aware and view-independent representations by randomly sampling the camera poses during training~\cite{KSchwarzNeurIPS2020,EChanCVPR2021}.
To prevent sampled camera parameters beyond a real distribution, we restrict its range (using a standard deviation of 0.1 radian in practice).
We found that this setting works reasonably well for both datasets, including limited and wide viewpoints (Section~\ref{subsec:ablation_study}).

\smallskip\noindent\textbf{Aperture ray sampling scheme.}
In typical ray tracing used in computer graphics~\cite{Shirley2020RTW1}, a large number of rays (e.g., 100) are sampled per pixel in aperture rendering (Section~\ref{subsec:aperture_rendering}) to improve the synthesis fidelity.
However, this increases both the processing time and memory.
To efficiently represent the aperture using limited rays, we used stratified sampling~\cite{BMildenhallECCV2020}.
More concretely, we used five rays; the origin of one ray was placed at the center of the aperture, and the origins of the others were placed along the circumference of the aperture with equal intervals.
We examine the effect of this approximation in Appendix~\ref{subsec:effect_sampling}.

\section{Experiments}
\label{sec:experiments}

\subsection{Experimental settings}
\label{subsec:experimental_settings}

We conducted two experiments to verify the effectiveness of AR-NeRF from multiple perspectives: a comparative study (Section~\ref{subsec:comparative_study}) in which we compared AR-NeRF to AR-GAN~\cite{TKanekoCVPR2021b}, which is a pioneering model with a similar objective, and an ablation study (Section~\ref{subsec:ablation_study}) in which we investigated the importance of our ideas.
In this section, we present the common settings and discuss the details of each in the next sections.

\smallskip\noindent\textbf{Dataset.}
Following the AR-GAN study~\cite{TKanekoCVPR2021b}, we evaluated AR-NeRF using three natural image datasets: two view-limited datasets, i.e., Oxford Flowers~\cite{MENilsbackICVGIP2008} (8,189 flower images with 102 categories) and CUB-200-2011~\cite{CWahCUB2002011} (11,788 bird images with 200 categories), and a view-various dataset, that is, FFHQ~\cite{TKarrasCVPR2019} (70,000 face images).
To effectively examine various cases, we resized the images to a pixel resolution of $64 \times 64$.
This strategy was also used in the AR-GAN study~\cite{TKanekoCVPR2021b}.
Therefore, we can compare AR-NeRF to AR-GAN under fair conditions.
We provide detailed information about the datasets in Appendix~\ref{subsec:implementation_main}.

\smallskip\noindent\textbf{Evaluation metrics.}
We evaluated the effectiveness of AR-NeRF quantitatively using the same two metrics used in the AR-GAN study~\cite{TKanekoCVPR2021b} for a direct comparison.
The first is the \textit{kernel inception distance (KID)}~\cite{MBinkowskiICLR2018}, which measures the maximum mean discrepancy between real and generated images within the inception model~\cite{CSzegedyCVPR2016}.
We used the KID to evaluate the visual fidelity of the generated images.
We calculated this score using 20,000 generated images and all real images.
Based on our objective (i.e., training an unconditional model on unstructured natural images), preparing the ground truth depth is nontrivial.
Following~\cite{TKanekoCVPR2021b}, as an alternative, we calculated the depth accuracy by (1) training the depth predictor using pairs of images and depths generated through GANs, (2) predicting the depths of real images using the trained depth predictor, and (3) comparing the predicted depths to those predicted by a highly generalizable monocular depth estimator~\cite{KXianCVPR2020} trained using stereo pairs.\footnote{We used the official pretrained model: \url{https://github.com/KexianHust/Structure-Guided-Ranking-Loss}.}
To measure the differences in depth, we used the \textit{scale-invariant depth error (SIDE)}~\cite{DEigenNIPS2014}, which measures the difference between depths in a scale-invariant manner and is useful for comparing the depths predicted by different representation systems.
For both metrics, the smaller the value, the better the performance.

\smallskip\noindent\textbf{Implementation.}
We implemented AR-NeRF based on pi-GAN~\cite{EChanCVPR2021},\footnote{We implemented it based on the official code: \url{https://github.com/marcoamonteiro/pi-GAN}.} which is a state-of-the-art generative variant of NeRF.
Because the original pi-GAN was not applied to the dataset used in our experiments, we carefully tuned the configurations and hyperparameters such that the baseline pi-GAN could generate images reasonably well.
Next, we incorporated a background synthesis network into pi-GAN based on NeRF++~\cite{KZhangArXiv2020}\footnote{We implemented this while referring to the official code: \url{https://github.com/Kai-46/nerfplusplus}.} (Section~\ref{subsec:advanced_techniques}).
Hereafter, we refer to this model as \textit{pi-GAN++}.
Subsequently, we incorporated aperture rendering (Section~\ref{subsec:aperture_rendering}) and aperture randomized training (Section~\ref{subsec:aperture_randomized_training}) into \textit{pi-GAN++}.
This is the model denoted by \textit{AR-NeRF} below.
We provide implementation details in Appendix~\ref{subsec:implementation_main}.

\subsection{Comparative study}
\label{subsec:comparative_study}

\begin{figure*}[t]
  \centering
  \includegraphics[width=0.9\textwidth]{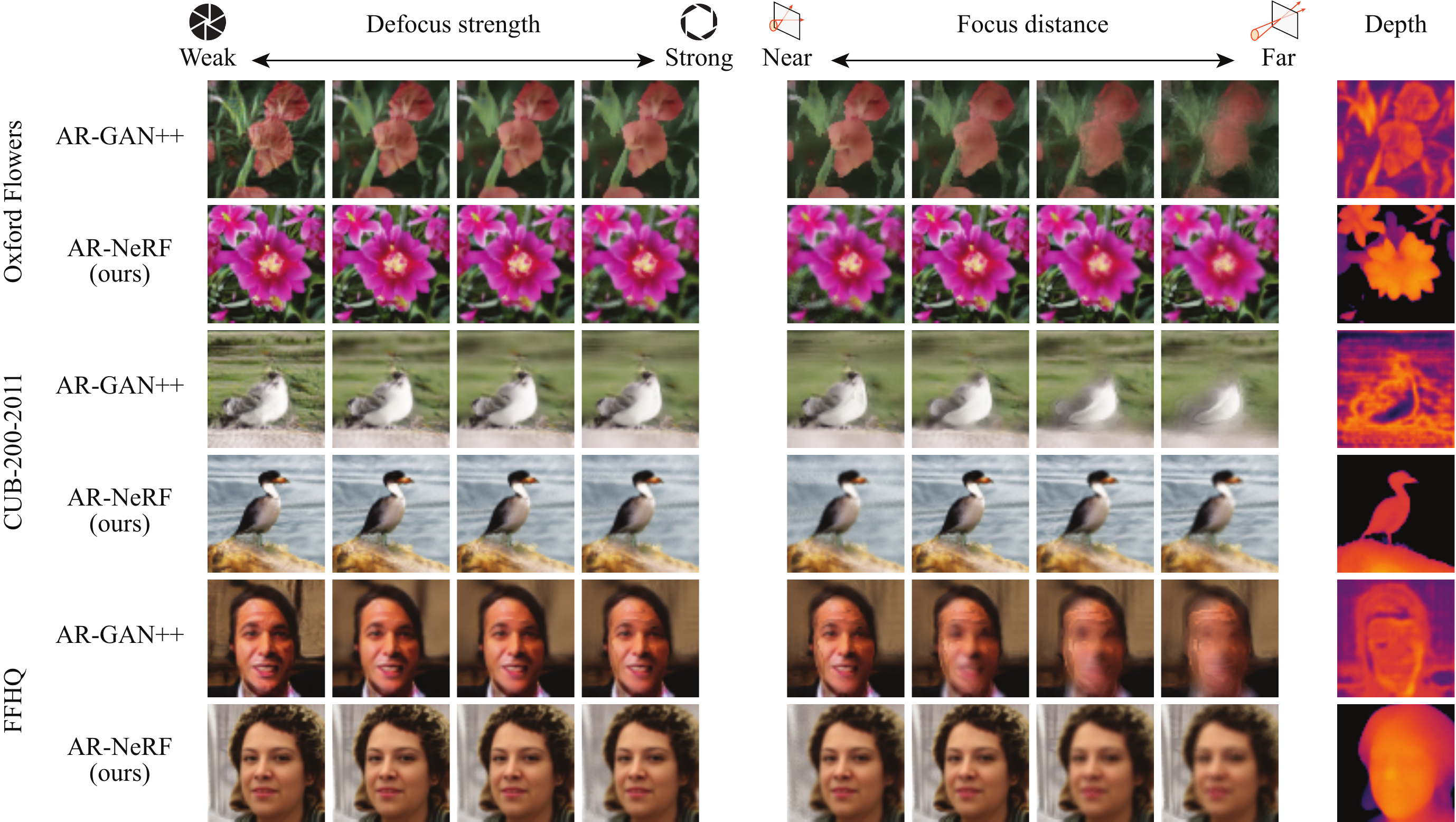}
  \vspace{-2mm}
  \caption{\textbf{Comparison of generated images and depths between AR-GAN++ and AR-NeRF (ours).}
    To manipulate the defocus strength, we varied the strength within $[0, \sigma_s, 2\sigma_s, 3\sigma_s]$, where $\sigma_s$ indicates the standard deviation of a half-normal distribution, which is used to represent the defocus distribution during training.
    To manipulate the focus distance, we used a range in which the foregrounds and backgrounds were focused.}
  \vspace{-4mm}
  \label{fig:generation}
\end{figure*}

To determine the validity of AR-NeRF for unsupervised learning of the \textit{depth} and \textit{defocus effects}, we first investigated the comparative performance between AR-NeRF and AR-GAN~\cite{TKanekoCVPR2021b}, which is a state-of-art model for this problem.
The main difference between AR-NeRF and AR-GAN is the architectural difference, where AR-NeRF is constructed based on the \textit{continuous} radiance fields, whereas AR-GAN is constructed based on \textit{discretized} CNNs.
Another small but significant difference is that AR-NeRF represents the defocus distribution (i.e., aperture size distribution) using a half-normal distribution (Equation~\ref{eqn:ar-nerf}), whereas AR-GAN represents it (i.e., depth scale distribution in this case) using a binomial distribution (Equation~6 in~\cite{TKanekoCVPR2021b}).
To confirm the effects of this difference, we also evaluated a variant of AR-GAN (referred to as \textit{AR-GAN++}), in which the defocus distribution was expressed using a half-normal distribution, similar to AR-NeRF.
Furthermore, as a reference, we report the scores of RGBD-GAN~\cite{ANoguchiICLR2020}, which learns the depth information using viewpoint cues.\footnote{In our preliminary experiments, we also examined the performances of AR-HoloGAN/AR-RGBD-GAN (combinations of AR-GAN and HoloGAN~\cite{TNguyenICCV2019}/RGBD-GAN~\cite{ANoguchiICLR2020}) on FFHQ.
  We found that the SIDE scores for AR-HoloGAN and AR-RGBD-GAN were 4.79 and 4.40, respectively, and were worse than those for AR-GAN.
  This result indicates that the simultaneous but individual usage of the viewpoint and defocus cues in AR-HoloGAN/AR-RGBD-GAN does not improve the depth learning.}

\begin{table}
  \centering
  \scriptsize
  \begin{tabularx}{\columnwidth}{cCCCCCC}
    \toprule
    & \multicolumn{2}{c}{Oxford Flowers}
    & \multicolumn{2}{c}{CUB-200-2011}
    & \multicolumn{2}{c}{FFHQ}
    \\ \cmidrule(r){2-3} \cmidrule(lr){4-5} \cmidrule(l){6-7}
    & KID$\downarrow$ & SIDE$\downarrow$
    & KID$\downarrow$ & SIDE$\downarrow$
    & KID$\downarrow$ & SIDE$\downarrow$
    \\ \midrule
    AR-GAN
    & 11.23 & 4.46
    & 14.30 & 3.58
    &  5.75 & 4.21
    \\
    AR-GAN++
    & 10.18 & 4.42
    & 13.91 & 3.61
    &  5.43 & 4.88
    \\
    RGBD-GAN
    & 12.04 & 7.01
    & 14.92 & 7.06
    &  6.73 & 5.81
    \\ \midrule
    AR-NeRF (ours)
    &  7.86 & 3.94
    &  6.81 & 3.63
    &  3.67 & 2.61
    \\ \bottomrule
  \end{tabularx}
  \vspace{-2mm}
  \caption{\textbf{Comparison of KID$\downarrow$ ($\times 10^3$) and SIDE$\downarrow$ ($\times 10^2$) between baseline GANs and AR-NeRF (ours).}}
  \label{tab:comparative_study}
  \vspace{-4mm}
\end{table}

\smallskip\noindent\textbf{Quantitative comparisons.}
We summarize quantitative comparison results in Table~\ref{tab:comparative_study}.
AR-NeRF outperformed the baseline GANs in terms of the KID and SIDE, except for SIDE on CUB-200-2011, where AR-GAN/AR-GAN++ was comparable to AR-NeRF.\footnote{SIDE has a limitation in that it can ignore certain types of degradation because it measures the difference based on $l_2$, causing statistical averaging.
  This may be why SIDE is comparable on CUB-200-2011 despite the qualitative differences (Figure~\ref{fig:generation}).
  To validate this hypothesis, we analyzed the gradient of the difference between the ground truth and predicted depths~\cite{DEigenICCV2015} and found that AR-NeRF outperformed AR-GAN/AR-GAN++ on this metric.
  We discuss the details in Appendix~\ref{subsec:delta_d}.}
These results validate the utility of AR-NeRF for unsupervised learning of the \textit{depth}.
We believe that the strengths of AR-NeRF, i.e., the joint usage of the viewpoint and defocus cues and continuous representations based on implicit functions, contribute to this improvement.
We present qualitative comparisons of the predicted depths in Figures~\ref{fig:depth_prediction_flowers}--\ref{fig:depth_prediction_faces} (Appendix~\ref{sec:additional_qualitative_results}).

\smallskip\noindent\textbf{Qualitative comparisons.}
We conducted qualitative comparisons to validate the effectiveness of unsupervised learning of the \textit{defocus effects}.
We present examples of generated images and depths in Figure~\ref{fig:generation}.
In AR-NeRF, we manipulated the defocus strength and focus distance by changing $s$ and $f$ (Figure~\ref{fig:ray_tracing}(b)).
As discussed above, the original AR-GAN discretely represents the defocus distribution.
Therefore, differently from AR-NeRF, it is unsuitable for conducting continuous operations.
Alternatively, we examine the performance of AR-GAN++, which represents a continuous defocus distribution.
In AR-GAN++, we manipulated the defocus strength and focus distance by changing the scale and offset of depth, respectively.

The results indicate that AR-NeRF can manipulate both the defocus strength and focus distance without generating significant artifacts.
In particular, in the manipulation of the focus distance, AR-NeRF succeeds in refocusing on both the foreground and background, the appearances of which are the same as those in the all-in-focus images (in the leftmost column).
By contrast, AR-GAN++ often generates unexpected artifacts, particularly when it attempts to refocus on the background (in the second-to-last column).
As possible causes for this phenomenon, 
(1) AR-GAN++ discretely represents light fields in a 2D space; thus, the discretization error becomes critical when large manipulations are counted and
(2) the predicted depths include artifacts (e.g., holes appearing in objects), causing errors when images are rendered based on depth.
The properties of AR-NeRF, that is, (1) a continuous representation in a 3D space and (2) joint usage of defocus and viewpoint cues, are useful for addressing these defects.
As another advantage of AR-NeRF, it can increase the resolution of the generated images by increasing the density of the input points owing to the nature of the implicit function~\cite{EChanCVPR2021}.
We demonstrate this strength in Figure~\ref{fig:concept}, where $128 \times 128$ images are generated using the same models as those used in Figure~\ref{fig:generation}.

\subsection{Ablation study}
\label{subsec:ablation_study}

We conducted ablation studies to examine the utility of AR-NeRF as a generative variant of the NeRF.
We compared AR-NeRF to five baselines:
\textit{pi-GAN}~\cite{EChanCVPR2021}, where a background synthesis network and aperture rendering are ablated;
\textit{pi-GAN++}, where aperture rendering is ablated;
\textit{AR-NeRF-0}, where viewpoint changes (Section~\ref{subsec:advanced_techniques}) are not applied during training;
\textit{AR-NeRF-F}, where the full viewpoint changes that are optimized to the face dataset (FFHQ) are used; and
\textit{pi-GAN++-F}, where aperture rendering is ablated from \textit{AR-NeRF-F}.
We tested the last two models on FFHQ only because viewpoint cues were limited on the other datasets.
For \textit{pi-GAN++}, we set the number of rays to be the same as that in AR-NeRF by an ensemble of multiple rays with an aperture size of $s = 0$.
We used this implementation to investigate the pure performance differences between the models with and without aperture rendering.

\begin{table}
  \centering
  \setlength{\tabcolsep}{1pt}
  \scriptsize
  \begin{tabularx}{\columnwidth}{ccccCCCCCC}
    \toprule
    & & &    
    & \multicolumn{2}{c}{Oxford Flowers}
    & \multicolumn{2}{c}{CUB-200-2011}
    & \multicolumn{2}{c}{FFHQ}
    \\ \cmidrule(){2-4} \cmidrule(lr){5-6} \cmidrule(lr){7-8} \cmidrule(lr){9-10}
    & (B) & (D) & (V)
    & KID$\downarrow$ & SIDE$\downarrow$
    & KID$\downarrow$ & SIDE$\downarrow$
    & KID$\downarrow$ & SIDE$\downarrow$
    \\ \midrule
    pi-GAN
    & & & L
    &  3.69 & 5.23
    &  5.04 & 4.87
    &  4.29 & 3.03
    \\
    pi-GAN++
    & \checkmark & & L
    &  8.30 & 4.83
    &  9.84 & 3.88
    &  4.43 & 2.69
    \\
    AR-NeRF-0
    & \checkmark & \checkmark & 0
    &  6.81 & 4.03
    &  8.67 & 3.74
    &  3.83 & 3.61
    \\
    AR-NeRF-F
    & \checkmark & \checkmark & F
    & -- & --
    & -- & --
    &  4.59 & 2.75
    \\
    pi-GAN++-F
    & \checkmark & & F
    & -- & --
    & -- & --
    &  5.06 & 2.78
    \\ \midrule
    AR-NeRF
    & \checkmark & \checkmark & L
    &  7.86 & 3.94
    &  6.81 & 3.63
    &  3.67 & 2.61
    \\ \bottomrule
  \end{tabularx}
  \vspace{-2mm}
  \caption{\textbf{Comparison of KID$\downarrow$ ($\times 10^3$) and SIDE$\downarrow$ ($\times 10^2$) between AR-NeRF and ablated models.}
    Check marks (B) and (D) indicate the use of a background synthesis network and defocus cue, respectively.
    In column (V), L, F, and 0 indicate the use of local, full, and no viewpoint changes, respectively.}
  \label{tab:ablation_study}
  \vspace{-4mm}
\end{table}

\smallskip\noindent\textbf{Results.}
We list the quantitative results in Table~\ref{tab:ablation_study} and provide a qualitative comparison of the predicted depths in Figures~\ref{fig:depth_prediction_flowers}--\ref{fig:depth_prediction_faces} (Appendix~\ref{sec:additional_qualitative_results}).
Our findings are as follows:

\smallskip\noindent\textit{(1) Effects of the background synthesis network (pi-GAN vs. pi-GAN++).}
We found that pi-GAN outperforms pi-GAN++ in terms of the KID.
We consider that the compact representation of the pi-GAN is advantageous for efficiently learning 2D image distributions.
However, pi-GAN was outperformed by pi-GAN++ in terms of the SIDE.
This result indicates that pi-GAN is unsuitable for our aims (i.e., unsupervised learning of the depth) despite its ability to generate high-fidelity images.

\smallskip\noindent\textit{(2) Effects of aperture rendering (pi-GAN++ vs. AR-NeRF).}
We found that AR-NeRF outperformed pi-GAN++ on both metrics, except for SIDE on FFHQ, where pi-GAN++ was comparable to AR-NeRF.
The same tendency holds for the comparison between AR-NeRF-F and pi-GAN++–F.
This is because FFHQ includes sufficient viewpoint variations to leverage the viewpoint cues.
By contrast, Oxford Flowers and CUB-200-2011 do not contain them.
In this case, the defocus cues used in AR-NeRF contributed to an improvement.

\smallskip\noindent\textit{(3) Comparison between viewpoint and defocus cues (pi-GAN++ vs. AR-NeRF-0)}.
With these models, the defocus and viewpoint manipulations are ablated.
Therefore, we can analyze each effect by comparing them.
We found that, in FFHQ, pi-GAN++ outperformed AR-NeRF-0 in terms of SIDE, whereas in the other datasets, AR-NeRF-0 outperformed pi-GAN++.
This can be explained by differences in the availability of the viewpoint cues, as discussed in (2).

\smallskip\noindent\textit{(4) Comparison between local and full viewpoint changes (AR-NeRF vs. AR-NeRF-F)}.
We found that AR-NeRF outperformed AR-NeRF-F on both metrics.
This result indicates that we do not need to carefully tune the camera parameters for unsupervised depth learning.
The same tendency was observed in the comparison between pi-GAN++ and pi-GAN++-F.
Note that AR-NeRF-F has an advantage in the viewpoint manipulation capability because it can learn full view variations, whereas AR-NeRF can only learn local view variations.

\smallskip\noindent\textbf{Detailed analyses.}
For further analyses, we examined (1) the importance of learning defocus effects from images (Appendix~\ref{subsec:importance_learning}), (2) the effect of the aperture ray sampling scheme (Appendix~\ref{subsec:effect_sampling}), (3) simultaneous control of the viewpoint and defocus (Appendix~\ref{subsec:control_viewpoint_defocus}), (4) generation of higher-resolution images (Appendix~\ref{subsec:generation_higher_resolution}), (5) application to defocus renderer (Appendix~\ref{subsec:application}), (6) the Fr\'{e}chet inception distance (FID)~\cite{MHeuselNIPS2017} (Appendix~\ref{subsec:fid}), and (7) the gradient of the difference in depth~\cite{DEigenICCV2015} (Appendix~\ref{subsec:delta_d}).
See the Appendices for further details.

\section{Discussion}
\label{sec:discussion}

\subsection{Limitations and future work}
\label{subsec:limitations}

AR-NeRF has two limitations, stemming from radiance field representations and fully unsupervised learning.

\smallskip\noindent\textbf{Limitations caused by radiance field representations.}
In radiance field representations, the computational complexity increases not only with the image size but also with the depth along each ray.
Consequently, the calculation cost is higher than that of a CNN GAN (e.g., AR-GAN~\cite{TKanekoCVPR2021b}); therefore, application to high-resolution images is difficult.
AR-NeRF requires multiple rays per pixel to represent aperture rendering. Therefore, it incurs a larger calculation cost than the standard NeRF, which represents each pixel using a single ray.
Through our experiments, we found that AR-NeRF outperforms the baseline NeRF, which has a similar calculation cost (in particular, pi-GAN++).
This demonstrates the validity of our research direction.
However, improving the calculation cost remains an important topic for future research.
Recent concurrent studies~\cite{JGuICLR2022,DLindellCVPR2021,CReiserICCV2021,AYuICCV2021,VSitzmannNeurIPS2021,SGarbinICCV2021} have addressed reducing the calculation cost of NeRF, and the incorporation of these methods is also a promising research area.

\begin{figure}[t]
  \centering
  \includegraphics[width=0.44\textwidth]{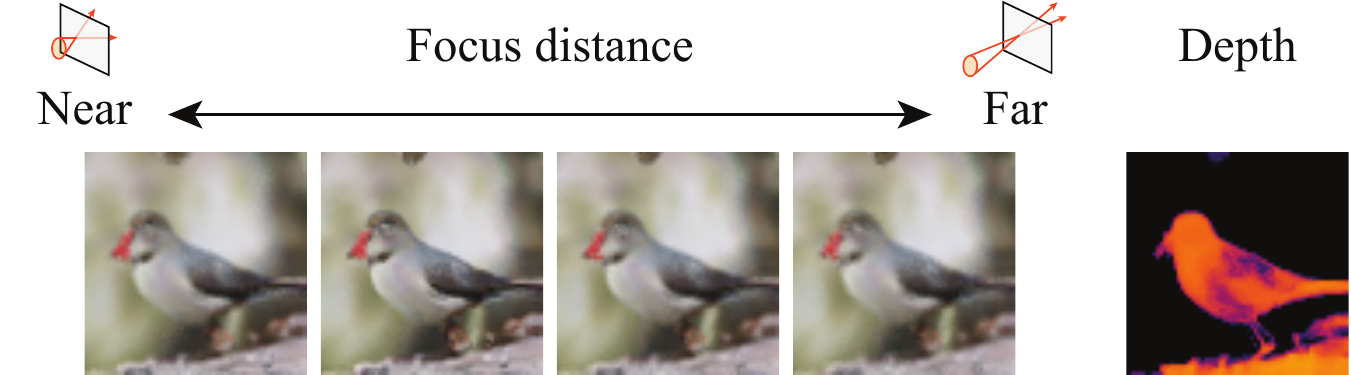}
  \vspace{-2mm}
  \caption{\textbf{Failure case.}}
  \vspace{-5mm}
  \label{fig:failure_case}
\end{figure}

\smallskip\noindent\textbf{Limitations caused by fully unsupervised learning.}
Fully unsupervised learning of the depth and defocus effects is highly challenging, and some limitations remain.
During our experiments, we found that our model is better than or comparable to the models trained under the same conditions.
However, its performance is lower than that of supervised models.
In particular, application to complex images will be difficult because AR-NeRF is a generative approach that assumes that it can learn the image generation reasonably well.
Furthermore, the use of an unbounded background based on NeRF++~\cite{KZhangArXiv2020} allows strong defocus effects that occur in the far plane to be represented.
However, it is still difficult to distinguish the defocus blur from the flat texture when the defocus blur is extremely strong (e.g., Figure~\ref{fig:failure_case}).
Addressing these problems is a possible direction for future research.

\subsection{Potential negative social impact}
\label{subsec:social_impact}

The method presented in this paper enables the creation of realistic images.
This poses a potential risk to the creation of misleading content (e.g., deepfake).
In particular, our model can increase the credibility of fake content in terms of 3D consistency and may potentially deceive systems that rely on 3D structures, such as face recognition systems.
Therefore, we believe that it is essential for the community to develop technology to distinguish fake images from real images and carefully monitor advancements in the corresponding research fields~\cite{SAgarwalCVPRW2019,FMarraMIPR2018,LNatarajEI2019,ARosslerICCV2019,YLiCVPR2020}.

\section{Conclusion}
\label{sec:conclusion}

To advance the research on the fully unsupervised learning of depth and defocus effects, we introduced AR-NeRF, which extends NeRF by incorporating aperture rendering.
AR-NeRF is noteworthy because it can employ defocus and viewpoint cues in a unified manner by representing both factors through a common ray-tracing framework.
We empirically demonstrated the effectiveness of AR-NeRF for unsupervised learning of the depth and defocus effects.
Although we focused on a generative variant of NeRF in this study, our idea, that is, the incorporation of aperture rendering in NeRF, is general, and we expect that its usage will broaden the applications of NeRF under practical scenarios.

\clearpage
{\small
  \bibliographystyle{ieee_fullname}
  \bibliography{egbib}
}

\clearpage
\appendix

\section*{Appendix}

This appendix provides detailed analyses (Appendix~\ref{sec:detailed_analyses}), additional qualitative results (Appendix~\ref{sec:additional_qualitative_results}), and implementation details (Appendix~\ref{sec:implementation_details}).

\section{Detailed analyses}
\label{sec:detailed_analyses}

In this section, we present seven detailed analyses to provide a deeper understanding of the proposed model.

\begin{itemize}
  \setlength{\parskip}{2pt}
  \setlength{\itemsep}{2pt}
\item Appendix~\ref{subsec:importance_learning}:
  Importance of learning defocus effects.
\item Appendix~\ref{subsec:effect_sampling}:
  Effect of aperture ray sampling scheme.
\item Appendix~\ref{subsec:control_viewpoint_defocus}:
  Simultaneous control of viewpoint and defocus.
\item Appendix~\ref{subsec:generation_higher_resolution}:
  Generation of higher-resolution images.
\item Appendix~\ref{subsec:application}:
  Application to defocus renderer.
\item Appendix~\ref{subsec:fid}:
  Fr\'{e}chet inception distance.
\item Appendix~\ref{subsec:delta_d}:
  Gradient of difference in depth.
\end{itemize}

\subsection{Importance of learning defocus effects}
\label{subsec:importance_learning}

As discussed in Section~\ref{subsec:aperture_rendering}, we represent aperture rendering in a ray-tracing framework, which is the basis of NeRF.
Thus, our idea is general; its application is not restricted to AR-NeRF and can be incorporated into any NeRF-based model, including existing or pretrained models.
This fact raises the question of whether it is necessary to learn the defocus effect from the data because we can attach the defocus effects virtually by incorporating aperture rendering into the model, even if aperture rendering is not used in the training.

In the main text, we present evidence verifying the importance of learning defocus effects from some perspectives.
More specifically, the results in Table~\ref{tab:ablation_study} indicate that the learning of the defocus effects is useful for improving model performance in terms of both image quality and depth accuracy.
In this section, we examine the tolerance to defocus manipulation.

When a model can manipulate the defocus strength to the greatest extent possible, we believe that it can generate an image that is close to a real image even when large defocus effects are imposed.
Based on this consideration, we examined the KID when large defocus effects were imposed on pi-GAN++ (i.e., a model without learning the defocus effects) and AR-NeRF (i.e., a model that learns the defocus effects).
More precisely, we calculated the KID when the aperture size $s \in \{ \sigma_s, 2\sigma_s, 3\sigma_s, 4\sigma_s, 5\sigma_s\}$, where $\sigma_s$ is the standard deviation of the aperture size obtained through AR-NeRF training.

\begin{table}[t]
  \centering
  \scriptsize
  \begin{tabularx}{\columnwidth}{cCCCCC}
    \toprule
    Oxford Flowers
    & $\sigma_s$
    & $2\sigma_s$
    & $3\sigma_s$
    & $4\sigma_s$
    & $5\sigma_s$
    \\ \midrule
    pi-GAN++
    & 8.27 & 10.63 & 19.40 & 32.04 & 43.94
    \\
    AR-NeRF
    & 7.86 & 9.07 & 14.21 & 22.09 & 31.91
    \\ \bottomrule
    \toprule
    CUB-200-2011
    & $\sigma_s$
    & $2\sigma_s$
    & $3\sigma_s$
    & $4\sigma_s$
    & $5\sigma_s$
    \\ \midrule
    pi-GAN++
    & 9.90 & 11.15 & 14.26 & 19.95 & 29.66
    \\
    AR-NeRF
    & 6.81 & 7.26 & 9.43 & 13.26 & 19.77
    \\ \bottomrule
    \toprule
    FFHQ
    & $\sigma_s$
    & $2\sigma_s$
    & $3\sigma_s$
    & $4\sigma_s$
    & $5\sigma_s$
    \\ \midrule
    pi-GAN++
    & 4.64 & 6.94 & 11.54 & 17.10 & 23.81
    \\
    AR-NeRF
    & 3.66 & 5.30 & 9.51 & 14.98 & 21.60
    \\ \bottomrule
  \end{tabularx}
  \vspace{-2mm}
  \caption{\textbf{Changes in the KID$\downarrow$ ($\times 10^3$) when varying the aperture size to $s \in \{ \sigma_s, 2\sigma_s, 3\sigma_s, 4\sigma_s, 5\sigma_s \}$.}
    $\sigma_s$ is the standard deviation of the aperture size obtained by training AR-NeRF.}
  \label{tab:aperture_effect}
  \vspace{-2mm}
\end{table}

\smallskip\noindent\textbf{Results.}
We summarize the results in Table~\ref{tab:aperture_effect}.
We found that for every dataset and aperture size, AR-NeRF outperformed pi-GAN++ in terms of the KID score.
In particular, we found that the difference is significant in the Oxford Flowers and CUB-200-2011 datasets, where viewpoint cues are limited or difficult to obtain and defocus cues play a critical role in obtaining a better SIDE (as shown in Table~\ref{tab:ablation_study}).
These results indicate that (1) neural radiance fields need to be optimized for defocus effects to obtain defocus-tolerant representations that can be jointly used in various ranges of defocus strength, and (2) AR-NeRF (i.e., a model that learns the defocus effects) is useful for addressing this problem, particularly when other cues (e.g., viewpoint cues) are limited.

\subsection{Effect of aperture ray sampling scheme}
\label{subsec:effect_sampling}

As discussed in Section~\ref{subsec:advanced_techniques}, stratified sampling ~\cite{BMildenhallECCV2020} was used to represent the aperture using a limited number of rays.
We used five rays in particular: the origin of one ray was placed at the center of the aperture, and the origins of the others were placed along the circumference of the aperture at equal intervals.
In this section, we examine the effect of this approximation.

Increasing the number of rays in the \textit{training} phase is costly; thus, we examined the effect of an aperture ray sampling scheme in the \textit{inference} phase.
More specifically, we compared \textit{stratified sampling}, which was used in the main experiments, with \textit{random sampling}, where the offsets of ray origins, that is, $\mathbf{u}$ in Equation~\ref{eqn:aperture_origin}, were randomly sampled in a disk of radius $s$ (i.e., $|\mathbf{u}| \in [0, s]$).
To examine the effect of the number of rays, we investigated the difference in performance when changing the number of rays in $\{ 1, 2, 5, 10, 20 \}$.
As mentioned above, we examined the difference in performance in an inference phase.
Hence, the base-trained model was the same as that used in Section~\ref{sec:experiments} and was common across all settings.

\smallskip\noindent\textbf{Results.}
We summarize the results in Table~\ref{tab:ray_effect}.
We found that, when we use random sampling,
(1) the performance is improved as the number of rays increases until reaching approximately 10,\footnote{The reason why the performance degrades when the number of rays increases too much (particularly in the cases where the number of rays is 20 on CUB-200-2011 and FFHQ) is that, in training, NeRF (including AR-NeRF) is optimized using \textit{finite} sample points.
  Consequently, \textit{overly dense} sampling in the inference phase can cause discrepancies from optimized conditions.
  This may lead to degradation when the number of rays increases too much.
  In other experiments, we found that the same phenomenon occurs when the number of sample points along the ray increases significantly.}
(2) the model with five rays (in the fourth column) performs worse than the model with the same number of rays with stratified sampling (in the last column), and
(3) we need to increase the number of rays to 10 (in the fifth column) to obtain a performance comparable to that of the model with stratified sampling (in the last column).
Considering that processing time and memory increase as the number of rays increases, we believe that stratified sampling with five rays is a reasonable choice in our experimental settings.

It is a possible that we will need to use more rays when applying to higher-resolution images, and in that case, we will need to increase the number of points along the ray.

\begin{table}[t]
  \centering
  \scriptsize
  \begin{tabularx}{\columnwidth}{cCCCCCC}
    \toprule
    Oxford Flowers
    & 1 & 2 & 5 & 10 & 20 & (5)
    \\ \midrule
    AR-NeRF
    & 18.80 & 12.17 & 8.51 & 7.65 & \textbf{7.37} & (7.86)
    \\ \bottomrule
    \toprule
    CUB-200-2011
    & 1 & 2 & 5 & 10 & 20 & (5)
    \\ \midrule
    AR-NeRF
    & 13.25 & 8.92 & 6.97 & \textbf{6.72} & 7.19 & (6.81)
    \\ \bottomrule
    \toprule
    FFHQ
    & 1 & 2 & 5 & 10 & 20 & (5)
    \\ \midrule
    AR-NeRF
    & 15.92 & 8.26 & 4.26 & \textbf{3.79} & 4.08 & (3.67)
    \\ \bottomrule
  \end{tabularx}
  \vspace{-2mm}
  \caption{\textbf{Changes in the KID$\downarrow$ ($\times 10^3$) when the number of rays varies within $\{ 1, 2, 5, 10, 20\}$.}
    When calculating the scores from the second to sixth columns, we randomly sampled the offsets of ray origins, that is, $\mathbf{u}$ (Equation~\ref{eqn:aperture_origin}), in a disk of radius $s$ (i.e., $|\mathbf{u}| \in [0, s]$).
    Bold font indicates the best scores.
    When calculating the scores in the last column, we used stratified sampling, as detailed in Section~\ref{subsec:advanced_techniques}.
    To distinguish them, we used parentheses for the latter.}
  \label{tab:ray_effect}
  \vspace{-2mm}
\end{table}

\subsection{Simultaneous control of viewpoint and defocus}
\label{subsec:control_viewpoint_defocus}

AR-NeRF is a natural extension of NeRF, in which pinhole camera-based ray tracing is replaced with aperture camera-based ray tracing.
This extension does not contradict the basic functionalities of NeRF. Thus, AR-NeRF can learn viewpoint-aware representations by taking over the characteristics of NeRF.

\begin{figure*}[p]
  \centering
  \includegraphics[height=0.95\textheight]{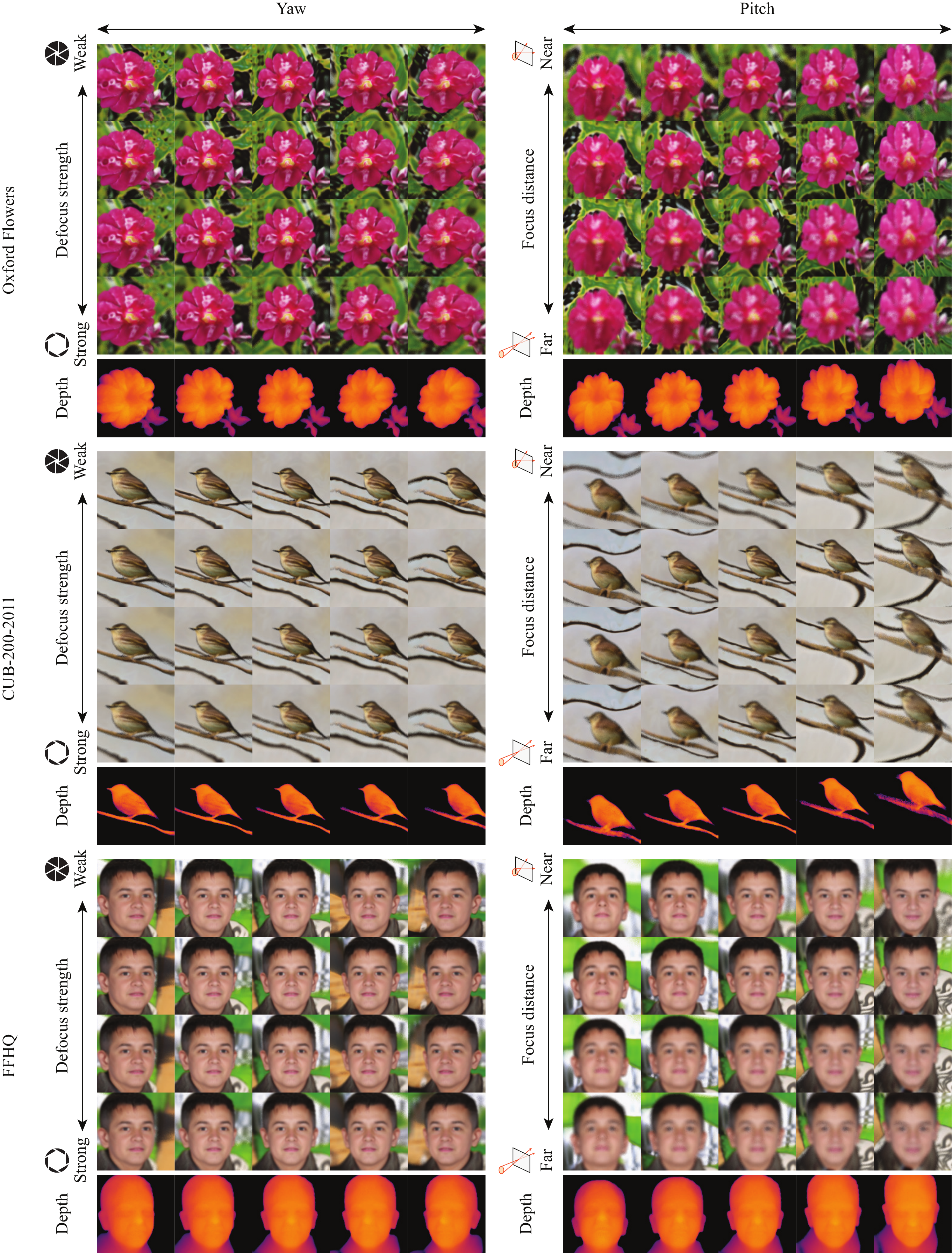}
  \caption{\textbf{Simultaneous control of viewpoint and defocus.}}
  \label{fig:control_viewpoint_defocus}
\end{figure*}

\smallskip\noindent\textbf{Results.}
We demonstrate this strength in Figure~\ref{fig:control_viewpoint_defocus}.
These results indicate that by using AR-NeRF, we can manipulate viewpoints and defocus effects simultaneously and independently in a unified framework.

\subsection{Generation of higher-resolution images}
\label{subsec:generation_higher_resolution}

In the experiments in the main text (Section~\ref{sec:experiments}), we used $64 \times 64$ images to examine various cases efficiently, following an AR-GAN study~\cite{TKanekoCVPR2021b}.
However, as discussed in Appendix~\ref{subsec:control_viewpoint_defocus}, AR-NeRF is a natural extension of NeRF; therefore, it can be applied to higher-resolution images with an increase in calculation cost, similar to other generative variants of NeRF, such as~\cite{KSchwarzNeurIPS2020,EChanCVPR2021}.
To validate this statement, we applied AR-NeRF to the $128 \times 128$ images.

\begin{figure*}[p]
  \centering
  \includegraphics[width=\textwidth]{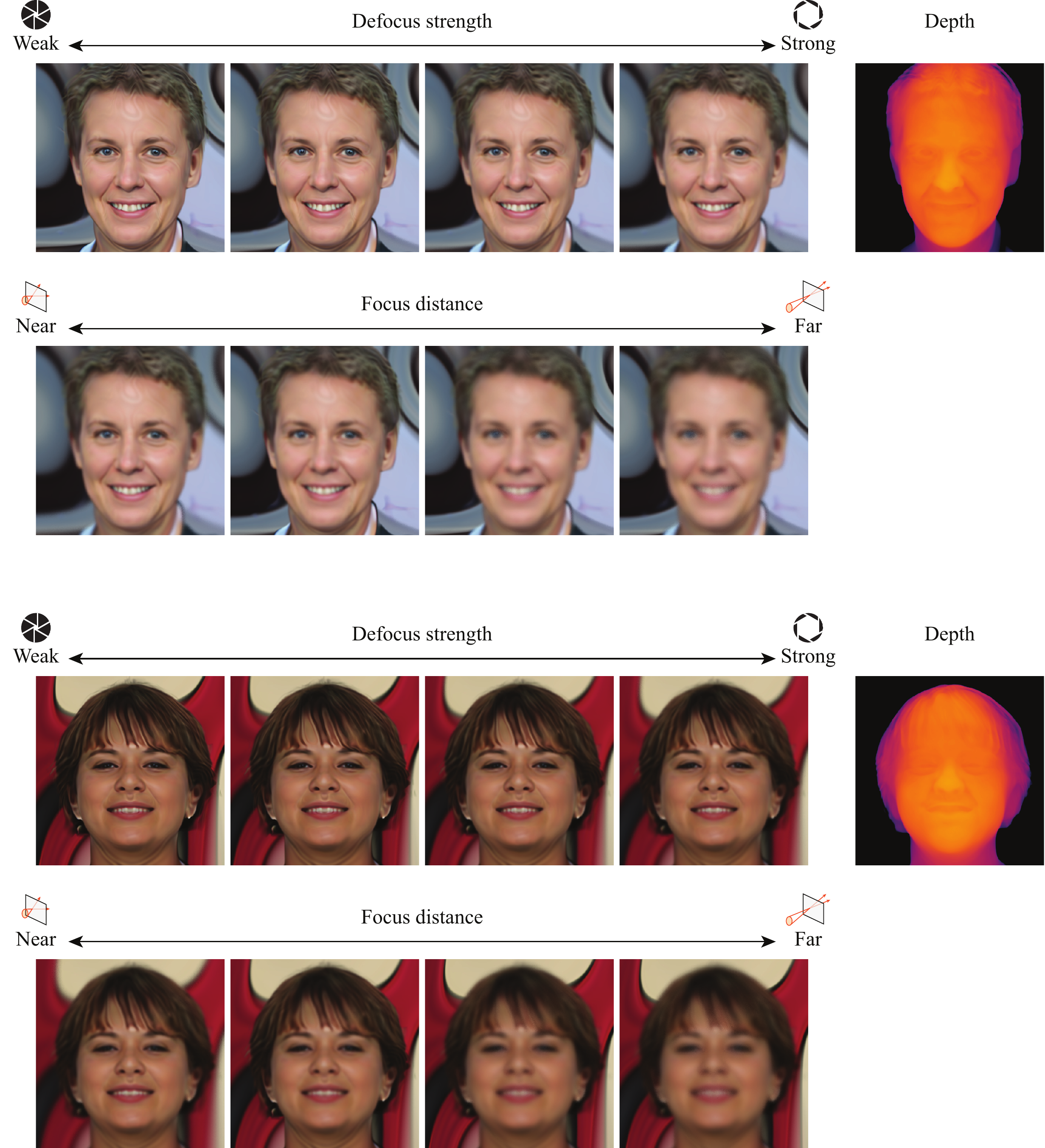}
  \caption{\textbf{Examples of $512 \times 512$ image and depth generation using AR-NeRF.}}
  \label{fig:generation_512}
\end{figure*}

\smallskip\noindent\textbf{Results.}
We provide the examples of generated images and depths in Figure~\ref{fig:generation_512}.
Following the experimental settings in the pi-GAN study~\cite{EChanCVPR2021}, we trained the model using $128 \times 128$ images and rendered the final results by sampling $512 \times 512$ pixels.
We found that AR-NeRF can render the defocus effects, including changes in defocus strength and focus distance, reasonably well, even in higher-resolution images.

\subsection{Application to defocus renderer}
\label{subsec:application}

After training, AR-NeRF can generate images from randomly sampled latent codes while varying the defocus strength and focus distance with photometric constraints.
By utilizing these images, we can train a defocus renderer, which, given an image, manipulates the defocus strength and focus distance intuitively and continuously.
We call this renderer \textit{AR-NeRF-R}.
In particular, we implemented AR-NeRF-R using a conditional extension of U-Net~\cite{ORonnebergerMICCAI2015,JYZhuNIPS2017}, which incorporates the aperture size $s$ and focus distance $f$ as auxiliary information to control the image generation based on them.
We provide the implementation details in Appendix~\ref{subsec:implementation_defocus_renderer}.
In this section, we empirically investigate the effectiveness of AR-NeRF-R.

\smallskip\noindent\textbf{Dataset.}
We used the Oxford Flowers dataset to train AR-NeRF and AR-NeRF-generated images to train AR-NeRF-R.
In particular, for AR-NeRF, we used the model discussed in Section~\ref{sec:experiments}.
When training AR-NeRF-R, we used $128 \times 128$ images generated by AR-NeRF, where we increased the resolution of the generated images from $64 \times 64$ to $128 \times 128$ by increasing the density of the input points (Section~\ref{subsec:comparative_study}), to allow AR-NeRF-R to be applied to $128 \times 128$ images.
To confirm the generality of the learned model, we evaluated it on a different dataset (\textit{iPhone2DSLR Flower}~\cite{JYZhuICCV2017}), including photographs of flowers taken by smartphones.
We provide the details regarding the dataset in Appendix~\ref{subsec:implementation_defocus_renderer}.

\smallskip\noindent\textbf{Comparison model.}
To the best of our knowledge, no previous method can learn the continuous representations of defocus strength and focus distance from natural images in the same setting as our own (i.e., \textit{without} any supervision and any predefined model).
Therefore, we used two baselines that have partially the same objective.
The first baseline is \textit{CycleGAN}~\cite{JYZhuICCV2017}, which trains a defocus renderer using \textit{set level} supervision.\footnote{We used the pretrained model provided by the authors: \url{https://github.com/junyanz/pytorch-CycleGAN-and-pix2pix}.}
In contrast to AR-NeRF-R, CycleGAN requires additional supervision to determine whether each training image is an all-in-focus or focused image.
The second baseline is \textit{AR-GAN-DR}~\cite{TKanekoCVPR2021b}, which can train a defocus renderer without any supervision or pretrained model, similar to AR-NeRF-R; however, its conversion is one-to-one, and it cannot adjust the defocus strength and focus distance continuously.

\begin{figure*}[p]
  \centering
  \includegraphics[width=\textwidth]{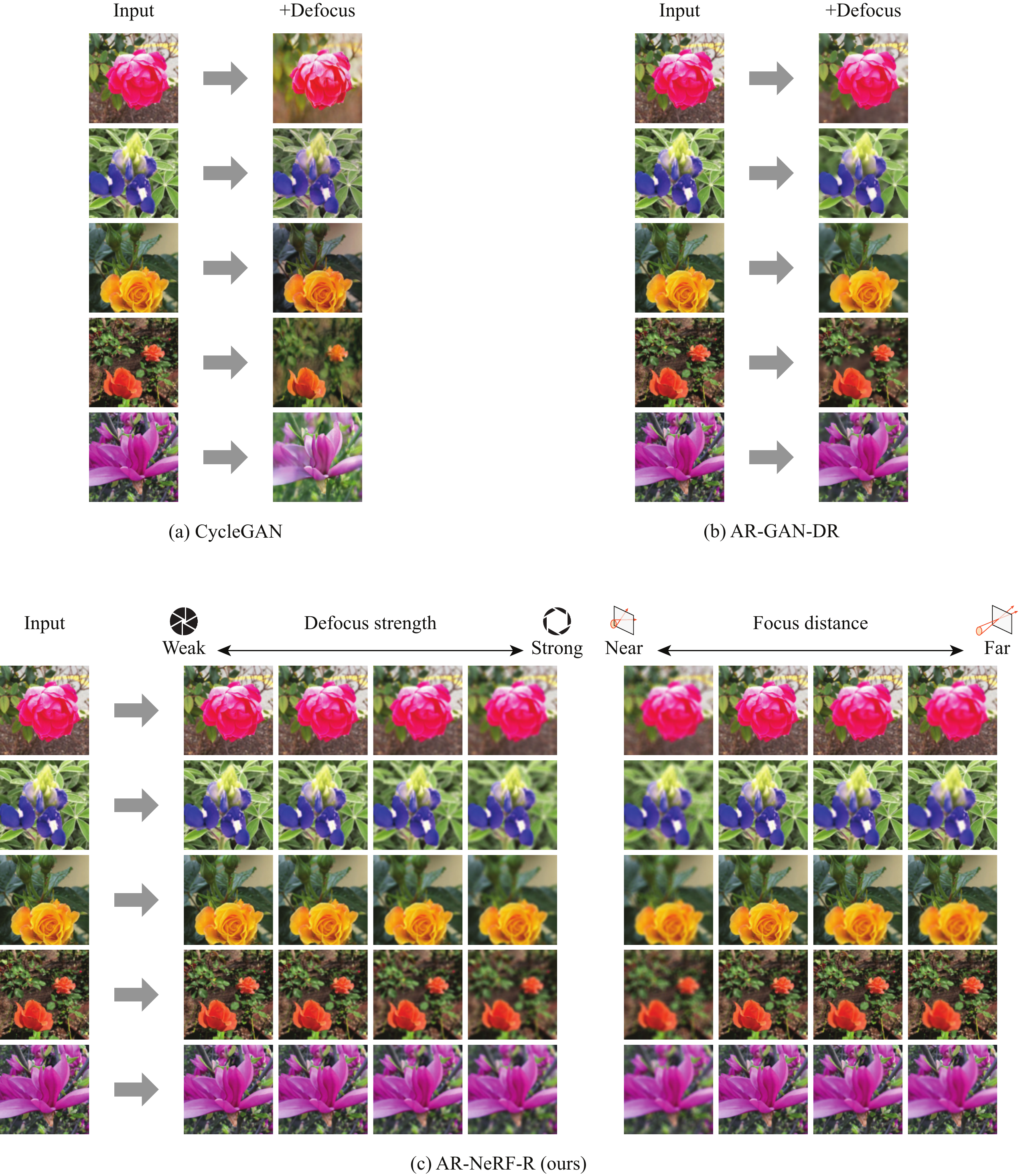}
  \caption{\textbf{Comparison of defocus rendering among CycleGAN, AR-GAN-DR, and AR-NeRF-R (ours).}}
  \label{fig:defocus_rendering}
\end{figure*}

\smallskip\noindent\textbf{Results.}
We present examples of the rendered images in Figure~\ref{fig:defocus_rendering}.
We found that CycleGAN often performs unnecessary changes (e.g., color changes in the fifth row), whereas AR-NeRF-R and AR-GAN-DR do not.
We infer that the aperture-rendering mechanisms in AR-NeRF and AR-GAN contributed to this phenomenon.
The difference between AR-GAN-DR and AR-NeRF-R is that in AR-GAN-DR, the defocus strength and focus distance are uniquely determined according to the input image, whereas in AR-NeRF-R, we can change them continuously by varying the auxiliary information of aperture size $s$ and focus distance $f$.
This new AR-NeRF-R functionality allows for interactive selection of defocused images.

\subsection{Fr\'{e}chet inception distance}
\label{subsec:fid}

In the main text, we used KID because it has an unbiased estimator and complements the flaws of other representative metrics (i.e., Fr\'{e}chet inception distance (FID)~\cite{MHeuselNIPS2017} and inception score (IS)~\cite{MHeuselNIPS2017}).
However, the FID is a widely used metric.
For reference, we report the FID in Tables~\ref{tab:comparative_study_extension} and~\ref{tab:ablation_study_extension}.
As also shown in a previous study~\cite{KKurachICML2019}, we found that KID and FID had high correlations in this case.
These results do not contradict the statements in the main text.

\subsection{Gradient of difference in depth}
\label{subsec:delta_d}

Figures~\ref{fig:depth_prediction_flowers}--\ref{fig:depth_prediction_faces} show that AR-GAN/AR-GAN++ yields unexpected artifacts around the edge and surface; however, SIDE can ignore this degradation because it measures the difference based on $l_2$, causing statistical averaging.
This may explain why the improvement in depth prediction by AR-NeRF is not reflected in SIDE on the CUB-200-2011 dataset (Table~\ref{tab:comparative_study}), despite the qualitative difference (Figure~\ref{fig:depth_prediction_birds}).
To validate this hypothesis, we calculated the gradient of the difference between the ground truth and predicted depths ($\nabla d$), which is commonly used to examine local structural similarity~\cite{DEigenICCV2015}.
Table~\ref{tab:comparative_study_extension2} lists the results and shows that AR-NeRF can improve depth prediction even on the CUB-200-2011 dataset in this metric.

\newpage
\begin{table}[h]
  \centering
  \scriptsize
  \begin{tabularx}{\columnwidth}{cCCCCCC}
    \toprule
    & \multicolumn{2}{c}{Oxford Flowers}
    & \multicolumn{2}{c}{CUB-200-2011}
    & \multicolumn{2}{c}{FFHQ}
    \\ \cmidrule(r){2-3} \cmidrule(lr){4-5} \cmidrule(l){6-7}
    & FID$\downarrow$ & KID$\downarrow$
    & FID$\downarrow$ & KID$\downarrow$ 
    & FID$\downarrow$ & KID$\downarrow$ 
    \\ \midrule
    AR-GAN
    & 20.4 & 11.23
    & 24.0 & 14.30
    & 10.4 &  5.75
    \\
    AR-GAN++
    & 19.0 & 10.18
    & 23.0 & 13.91
    & \hspace{4pt}9.9 &  5.43
    \\
    RGBD-GAN
    & 20.8 & 12.04
    & 24.6 & 14.92
    & 11.6 &  6.73
    \\ \midrule
    AR-NeRF
    & 17.1 & \hspace{3pt}7.86
    & 17.0 & \hspace{3pt}6.81
    & \hspace{4pt}7.8 &  3.67
    \\ \bottomrule
  \end{tabularx}
  \vspace{-2mm}
  \caption{\textbf{Comparison of FID$\downarrow$ and KID$\downarrow$ ($\times 10^3$) between baseline GANs and AR-NeRF (ours).}
    This table supplements Table~\ref{tab:comparative_study}.}
  \label{tab:comparative_study_extension}
  \vspace{-2mm}
\end{table}

\begin{table}[h]
  \centering
  \setlength{\tabcolsep}{1pt}
  \scriptsize
  \begin{tabularx}{\columnwidth}{ccccCCCCCC}
    \toprule
    & & &    
    & \multicolumn{2}{c}{Oxford Flowers}
    & \multicolumn{2}{c}{CUB-200-2011}
    & \multicolumn{2}{c}{FFHQ}
    \\ \cmidrule(){2-4} \cmidrule(lr){5-6} \cmidrule(lr){7-8} \cmidrule(lr){9-10}
    & (B) & (D) & (V)
    & FID$\downarrow$ & KID$\downarrow$
    & FID$\downarrow$ & KID$\downarrow$
    & FID$\downarrow$ & KID$\downarrow$
    \\ \midrule
    pi-GAN
    & & & L
    & 12.6 &  3.69
    & 14.8 &  5.04
    &  9.7 &  4.29
    \\
    pi-GAN++
    & \checkmark & & L
    & 18.2 &  8.30
    & 21.5 &  9.84
    &  8.6 &  4.43
    \\
    AR-NeRF-0
    & \checkmark & \checkmark & 0
    & 15.2 &  6.81
    & 20.1 &  8.67
    &  8.0 &  3.83
    \\
    AR-NeRF-F
    & \checkmark & \checkmark & F
    & -- & --
    & -- & --
    &  8.8 &  4.59
    \\
    pi-GAN++-F
    & \checkmark & & F
    & -- & --
    & -- & --
    &  9.8 &  5.06
    \\ \midrule
    AR-NeRF
    & \checkmark & \checkmark & L
    & 17.1 &  7.86
    & 17.0 &  6.81
    &  7.8 &  3.67
    \\ \bottomrule
  \end{tabularx}
  \vspace{-2mm}
  \caption{\textbf{Comparison of FID$\downarrow$ and KID$\downarrow$ ($\times 10^3$) between AR-NeRF and ablated models.}
    This table supplements Table~\ref{tab:ablation_study}.
    Check marks (B) and (D) indicate the use of a background synthesis network and defocus cue, respectively.
    In column (V), L, F, and 0 indicate the use of local, full, and no viewpoint changes, respectively.}
  \label{tab:ablation_study_extension}
  \vspace{-2mm}
\end{table}

\begin{table}[h]
  \setlength{\tabcolsep}{1pt}
  \centering
  \scriptsize
  \begin{tabularx}{\columnwidth}{cCCCCCCCCC}
    \toprule
    & \multicolumn{3}{c}{Oxford Flowers}
    & \multicolumn{3}{c}{CUB-200-2011}
    & \multicolumn{3}{c}{FFHQ}
    \\ \cmidrule(r){2-4} \cmidrule(lr){5-7} \cmidrule(l){8-10}
    & KID$\downarrow$ & SIDE$\downarrow$ & $\nabla d$$\downarrow$
    & KID$\downarrow$ & SIDE$\downarrow$ & $\nabla d$$\downarrow$
    & KID$\downarrow$ & SIDE$\downarrow$ & $\nabla d$$\downarrow$
    \\ \midrule
    AR-GAN
    & 11.23 & 4.46 & 6.94
    & 14.30 & 3.58 & 4.99
    &  5.75 & 4.21 & 5.73
    \\
    AR-GAN++
    & 10.18 & 4.42 & 7.01
    & 13.91 & 3.61 & 4.99
    &  5.43 & 4.88 & 7.37
    \\ \midrule
    AR-NeRF
    &  7.86 & 3.94 & 3.54
    &  6.81 & 3.63 & 3.39
    &  3.67 & 2.61 & 2.24
    \\ \bottomrule
  \end{tabularx}
  \vspace{-2mm}
  \caption{\textbf{Comparison of KID$\downarrow$ ($\times 10^3$), SIDE$\downarrow$ ($\times 10^2$), and $\nabla d$$\downarrow$ ($\times 10^2$) among AR-GAN, AR-GAN++, and AR-NeRF (ours).}
    This table supplements Table~\ref{tab:comparative_study}.}
  \label{tab:comparative_study_extension2}
  \vspace{-2mm}
\end{table}

\clearpage
\section{Additional qualitative results}
\label{sec:additional_qualitative_results}

\begin{figure*}[p]
  \centering
  \includegraphics[width=0.95\textwidth]{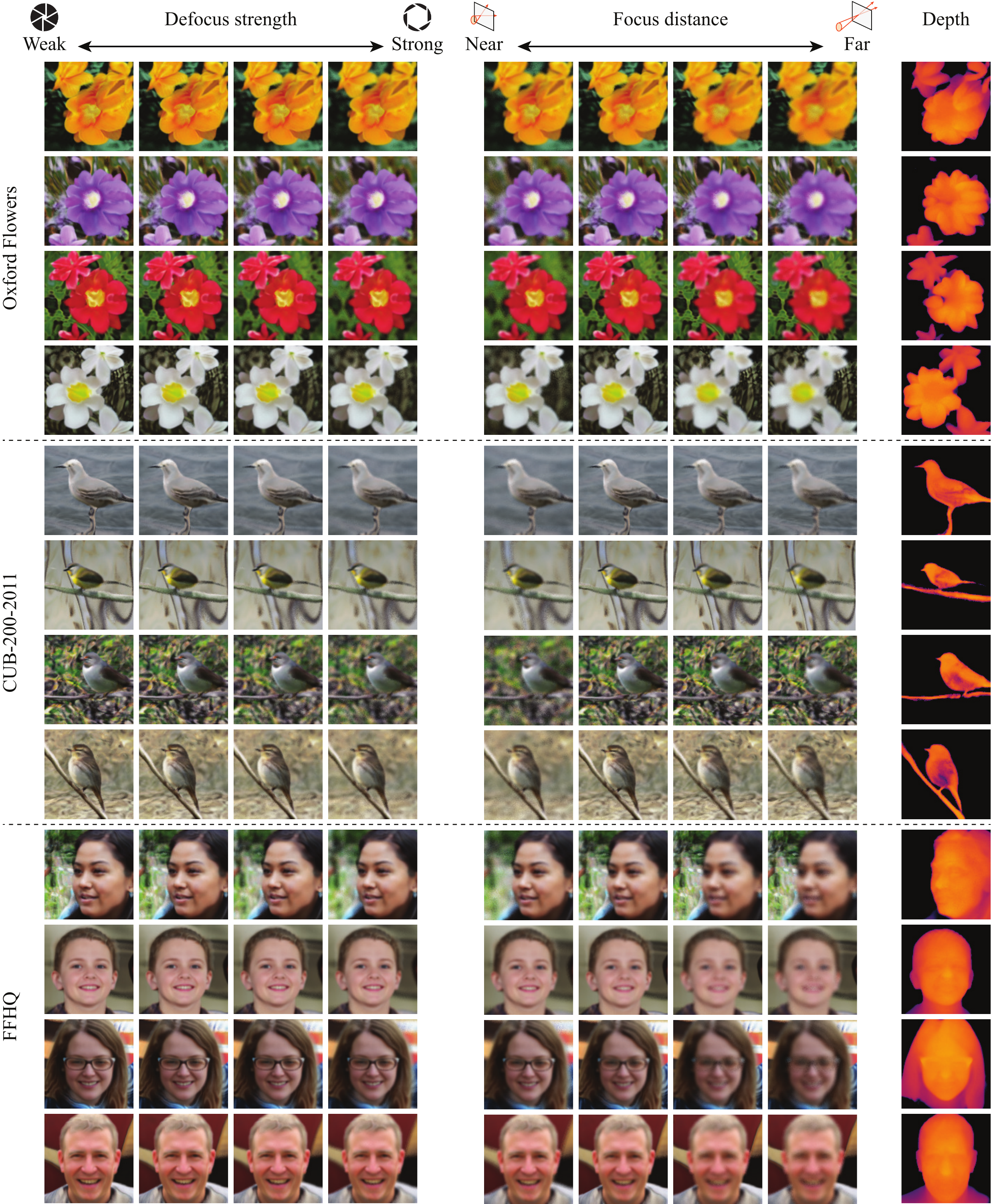}
  \caption{\textbf{Unsupervised learning of depth and defocus effects from unstructured (and view-limited) natural images.}
    This figure is an extended version of Figure~\ref{fig:concept}.
    As shown here, our objective is to acquire a generator that can generate sets of images and depths using only a collection of unstructured (and view-limited) single images and without any supervision (e.g., ground-truth depth, pairs of multiview images, defocus supervision, and pretrained models).
    In particular, in the generation of an image, we aim to obtain a generator that can intuitively and continuously adjust the defocus strength and focus distance with photometric constraints.}
  \label{fig:concept_extension}
\end{figure*}

\begin{figure*}[p]
  \centering
  \includegraphics[width=0.99\textwidth]{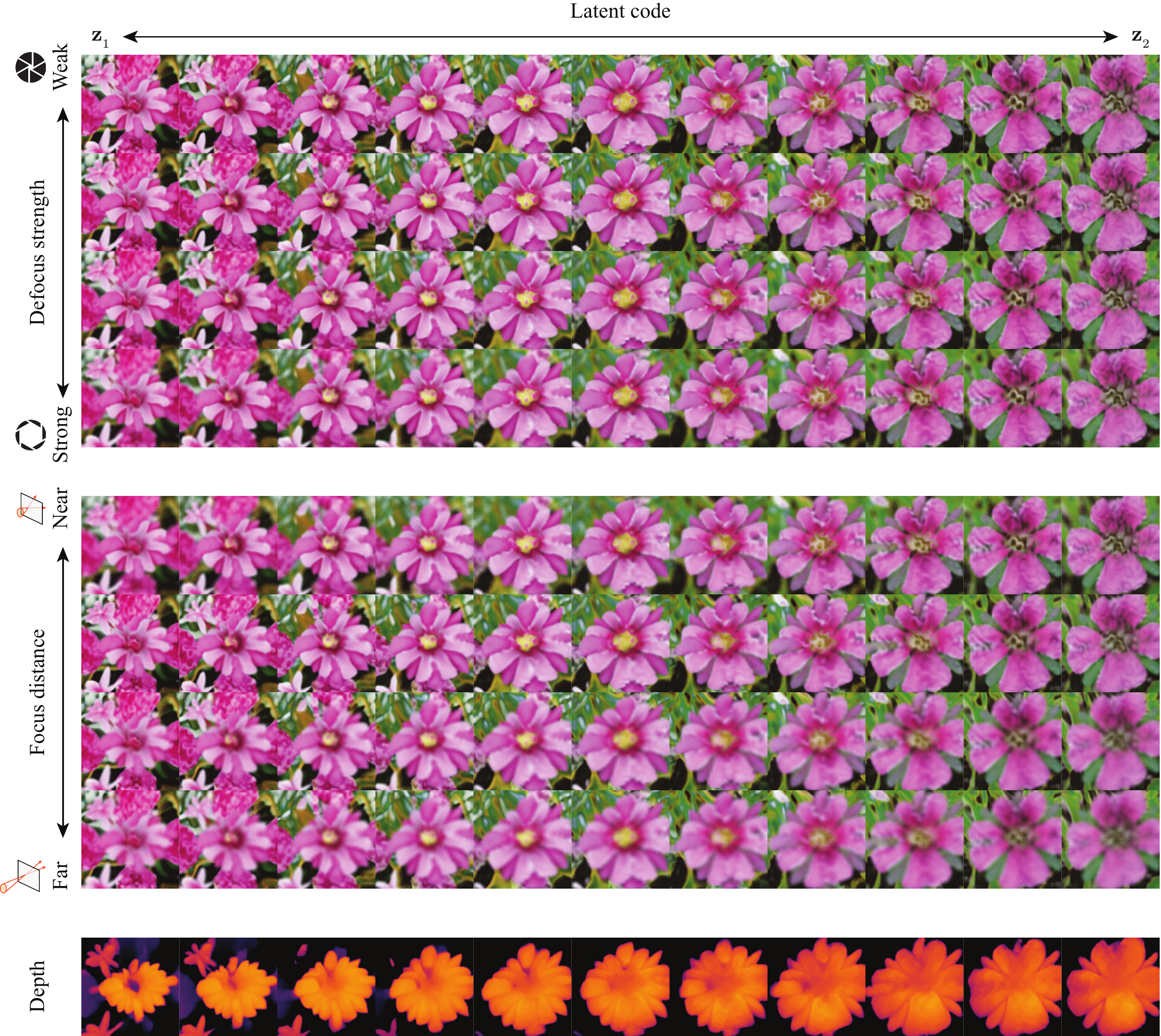}
  \caption{\textbf{Simultaneous control of defocus and latent codes on the Oxford Flowers dataset.}
    This figure is an extended version of Figure~\ref{fig:concept}.
    In AR-NeRF, aperture randomized training (Section~\ref{subsec:aperture_randomized_training}) encourages the defocus effects and latent codes to capture independent representations.
    By employing this characteristic, we can manipulate defocus effects and latent codes independently and simultaneously.}
  \label{fig:control_defocus_latent_flowers}
\end{figure*}

\begin{figure*}[p]
  \centering
  \includegraphics[width=0.99\textwidth]{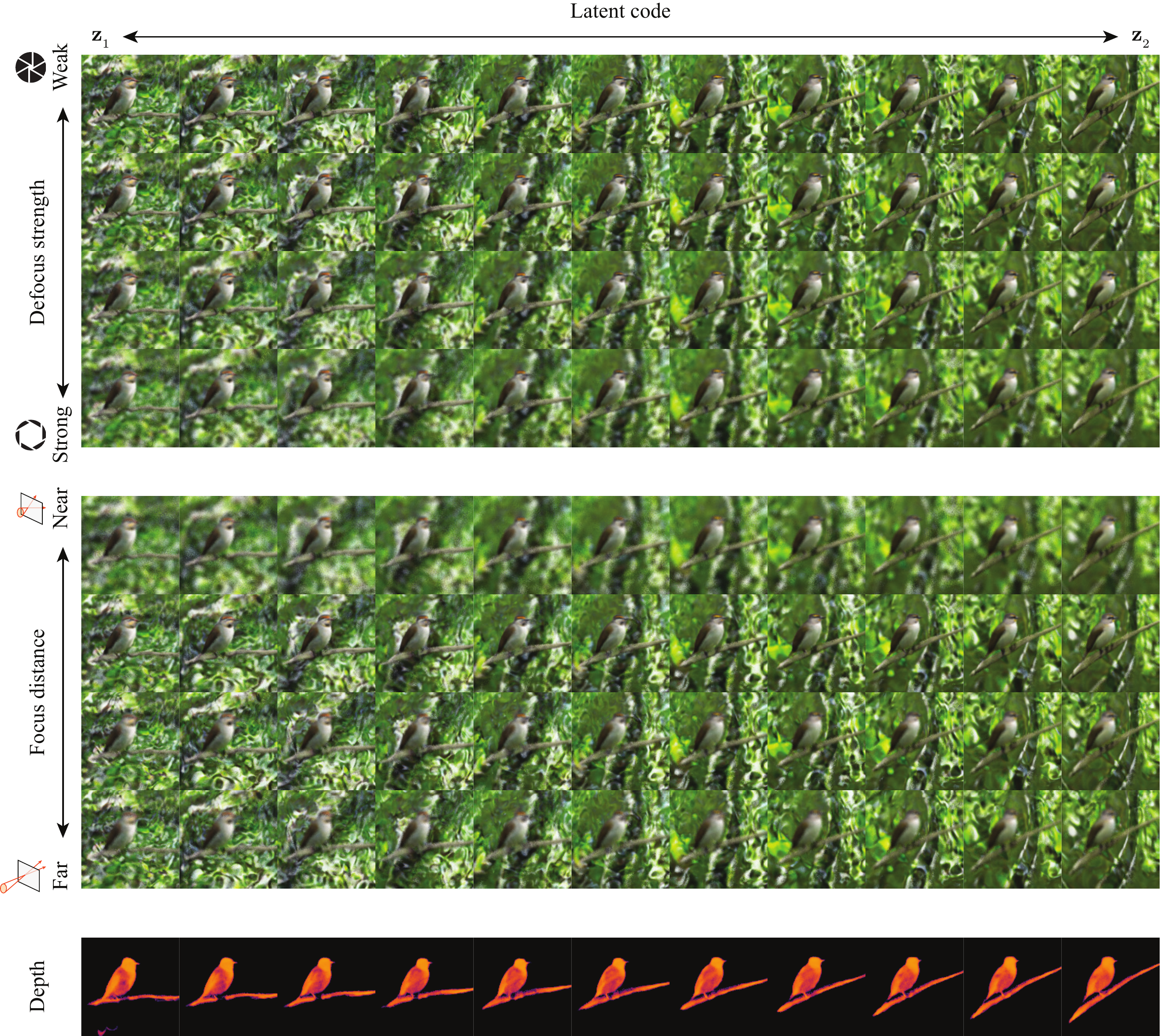}
  \caption{\textbf{Simultaneous control of defocus and latent codes on the CUB-200-2011 dataset.}
    This figure is an extended version of Figure~\ref{fig:concept}.
    In AR-NeRF, aperture randomized training (Section~\ref{subsec:aperture_randomized_training}) encourages the defocus effects and latent codes to capture independent representations.
    By employing this characteristic, we can manipulate defocus effects and latent codes independently and simultaneously.}
  \label{fig:control_defocus_latent_birds}
\end{figure*}

\begin{figure*}[p]
  \centering
  \includegraphics[width=0.99\textwidth]{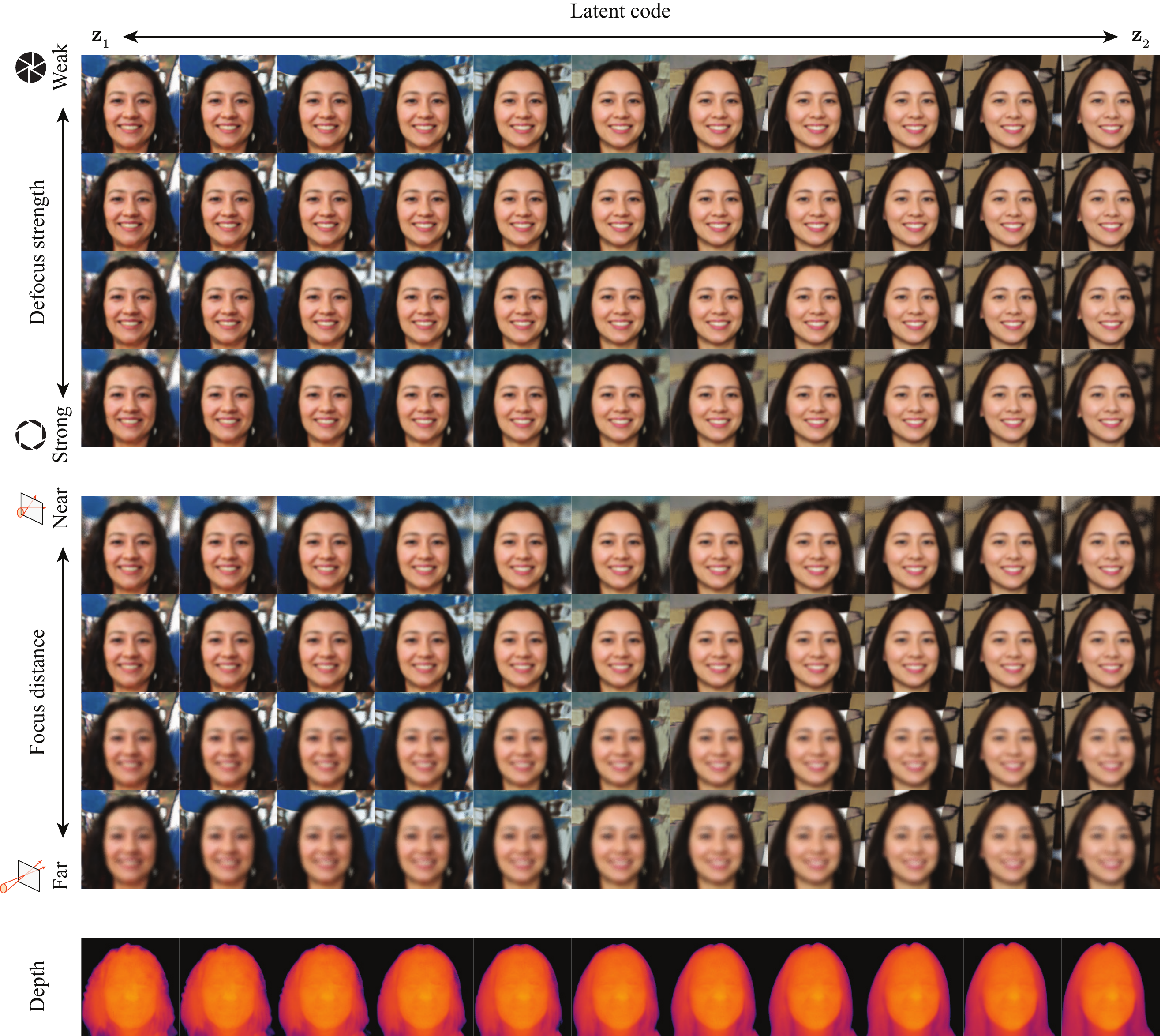}
  \caption{\textbf{Simultaneous control of defocus and latent codes on the FFHQ dataset.}
    This figure is an extended version of Figure~\ref{fig:concept}.
    In AR-NeRF, aperture randomized training (Section~\ref{subsec:aperture_randomized_training}) encourages the defocus effects and latent codes to capture independent representations.
    By employing this characteristic, we can manipulate defocus effects and latent codes independently and simultaneously.}
  \label{fig:control_defocus_latent_faces}
\end{figure*}

\begin{figure*}[p]
  \centering
  \includegraphics[width=0.93\textwidth]{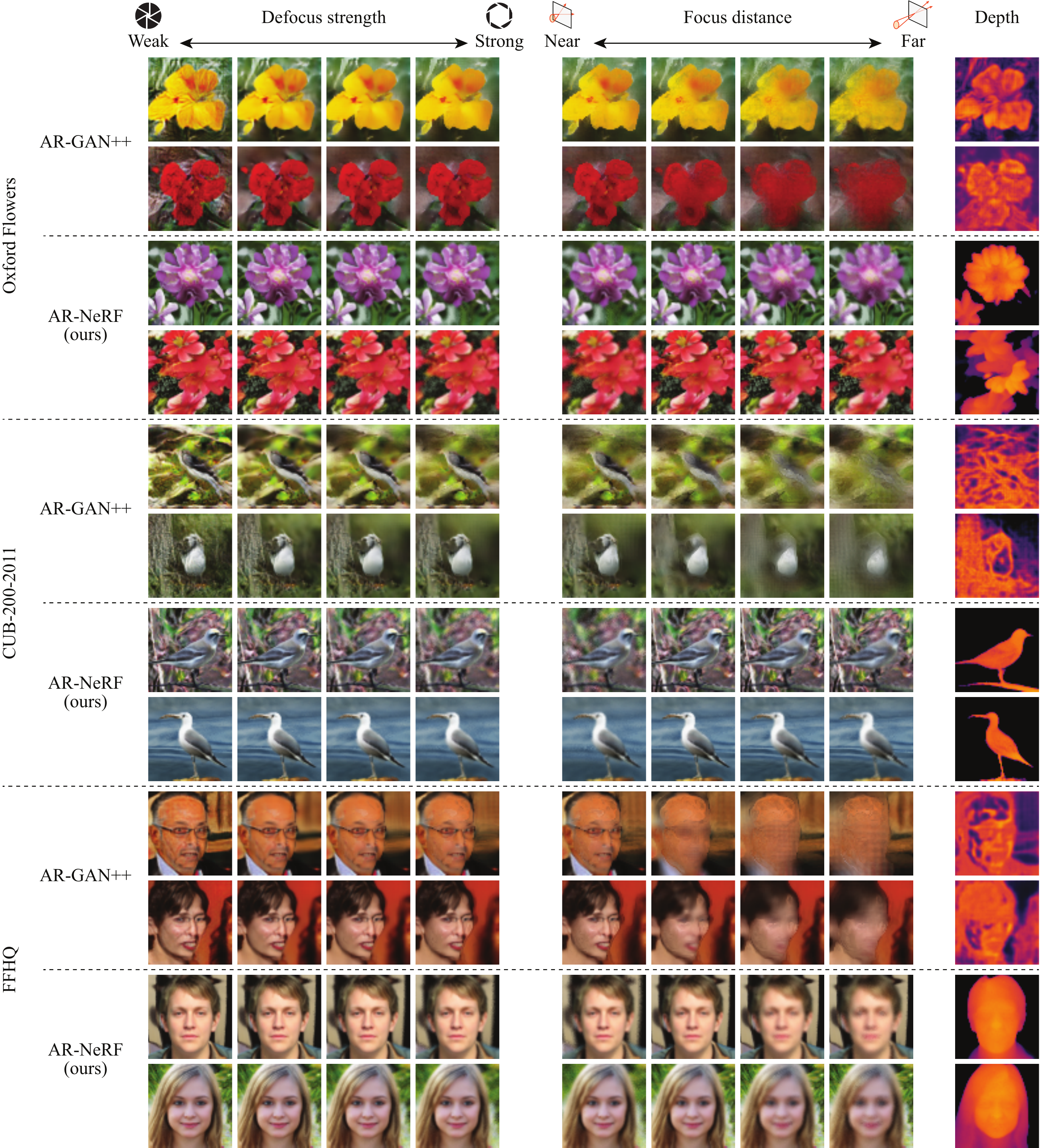}
  \caption{\textbf{Comparison of generated images and depths between AR-GAN++ and AR-NeRF (ours).}
    This figure is an extended version of Figure~\ref{fig:generation}.
    We found that AR-NeRF can manipulate both the defocus strength and focus distance without producing significant artifacts.
    In particular, it is worth noting that AR-NeRF can refocus on both the foreground (shown in the fifth column) and background (shown in the second-to-last column), which are almost the same as those in the all-in-focus images (shown in the first column), by manipulating the focus distance.
    By contrast, AR-GAN++ tends to yield unexpected artifacts (e.g., over-smoothing or discretization artifacts), particularly when there is a strong defocus (shown in the fourth column) or refocus on the background (shown in the second-to-last column).
    As discussed in the main text, the possible causes for these phenomena are: (1) AR-GAN++ discretely represents light fields in a 2D space; thus, the discretization error becomes critical when a large manipulation is performed, and (2) the predicted depths (shown in the last column) contain artifacts (e.g., holes emerge in the objects), resulting in errors when images are rendered based on the depths.
    The properties of AR-NeRF, that is, (1) continuous representation in a 3D space and (2) joint optimization using defocus and viewpoint cues, are useful for addressing these weaknesses.}
  \label{fig:generation_extension}
\end{figure*}

\begin{figure*}[p]
  \centering
  \includegraphics[width=0.8\textwidth]{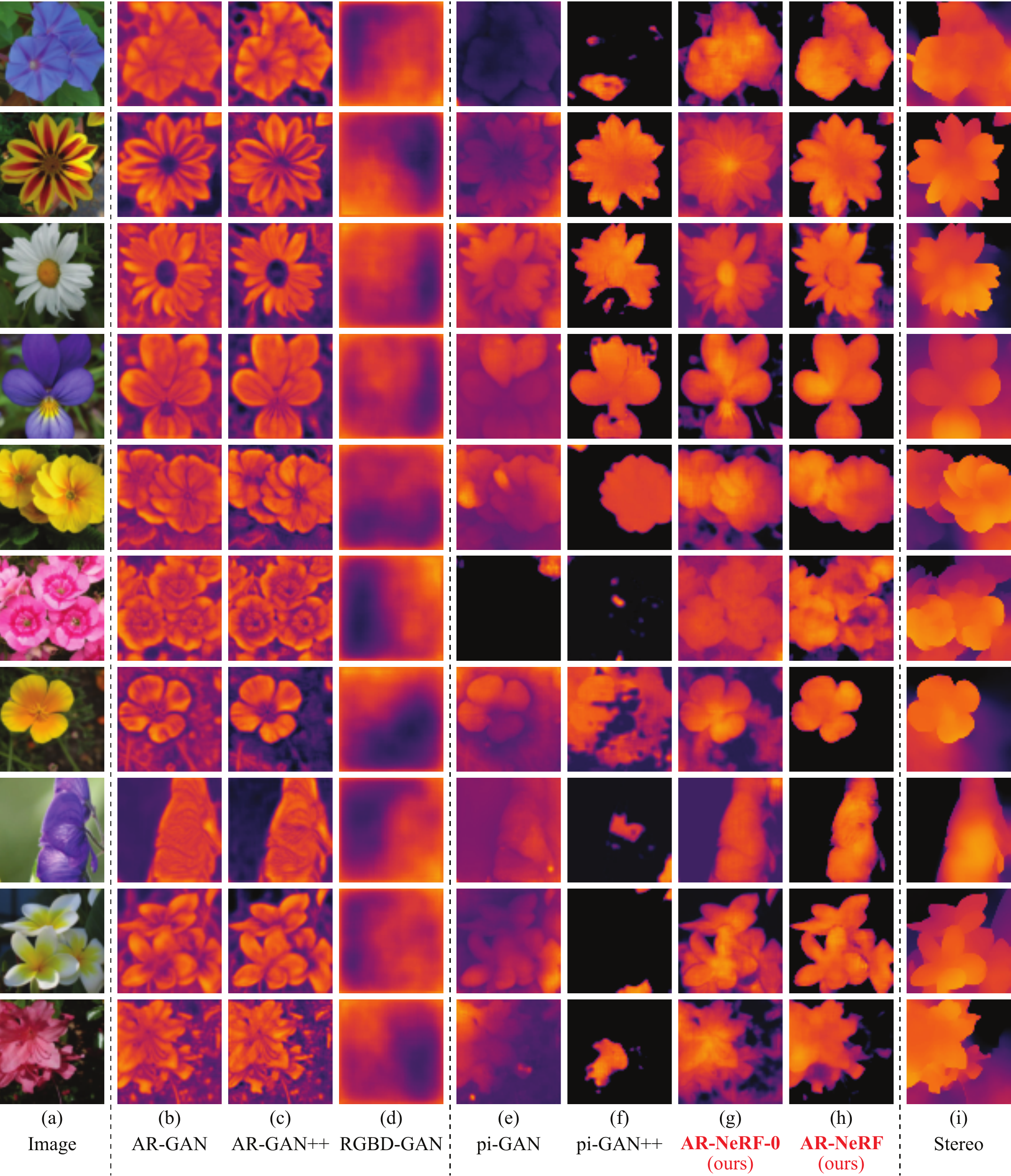}
  \caption{\textbf{Comparison of depth prediction on the Oxford Flowers dataset.}
    These depths are used to calculate the SIDEs in Tables~\ref{tab:comparative_study} and \ref{tab:ablation_study}.
    AR-GAN (b), AR-GAN++ (c), and RGBD-GAN (d) are CNN-based and are trained in a \textit{fully} unsupervised manner.
    In addition, pi-GAN (e), pi-GAN++ (f), AR-NeRF-0 (g), and AR-NeRF (h) are NeRF-based and trained in a \textit{fully} unsupervised manner.
    By contrast, the model in (i)~\cite{KXianCVPR2020} was trained using stereo supervision and applied as the \textit{ground truth} in the evaluation.
    The SIDEs in Tables~\ref{tab:comparative_study} and \ref{tab:ablation_study} were calculated by comparing the depths in (b)--(h) with the depths in (i).
    Our findings are summarized as follows:
    (1) Because AR-GAN (b) and AR-GAN++ (c) only employ defocus (appearance) cues, their predicted depths are affected by their appearance.
    For example, in the second row, the pattern in the petals affects depth prediction, despite its non-necessity.
    (2) RGBD-GAN (d) utilizes viewpoint (geometric) cues for 3D representation learning.
    However, there are few viewpoint cues in this dataset; consequently, this model has difficulty in learning depth.
    (3) For the same reason, pi-GAN (e) and pi-GAN++ (f), which only employ viewpoint cues, suffer from learning difficulties, although NeRF itself has a strong 3D consistency at the design level.
    In particular, they fail to consistently distinguish between the foreground (flower) and background (surroundings), parts of which are often missing.
    (4) Although the results of AR-NeRF-0 (g) are closest to those of AR-NeRF (h), AR-NeRF-0 (g) is often affected by the appearance (e.g., in the second row, similar to AR-GAN (b) and AR-GAN++ (c)) because it also leverages only focus cues.
    (5) AR-NeRF (h) overcomes the limitations of pi-GAN (e), pi-GAN++ (f), and AR-NeRF-0 (g) by utilizing both the viewpoint and defocus cues.}
  \label{fig:depth_prediction_flowers}
\end{figure*}

\begin{figure*}[p]
  \centering
  \includegraphics[width=0.8\textwidth]{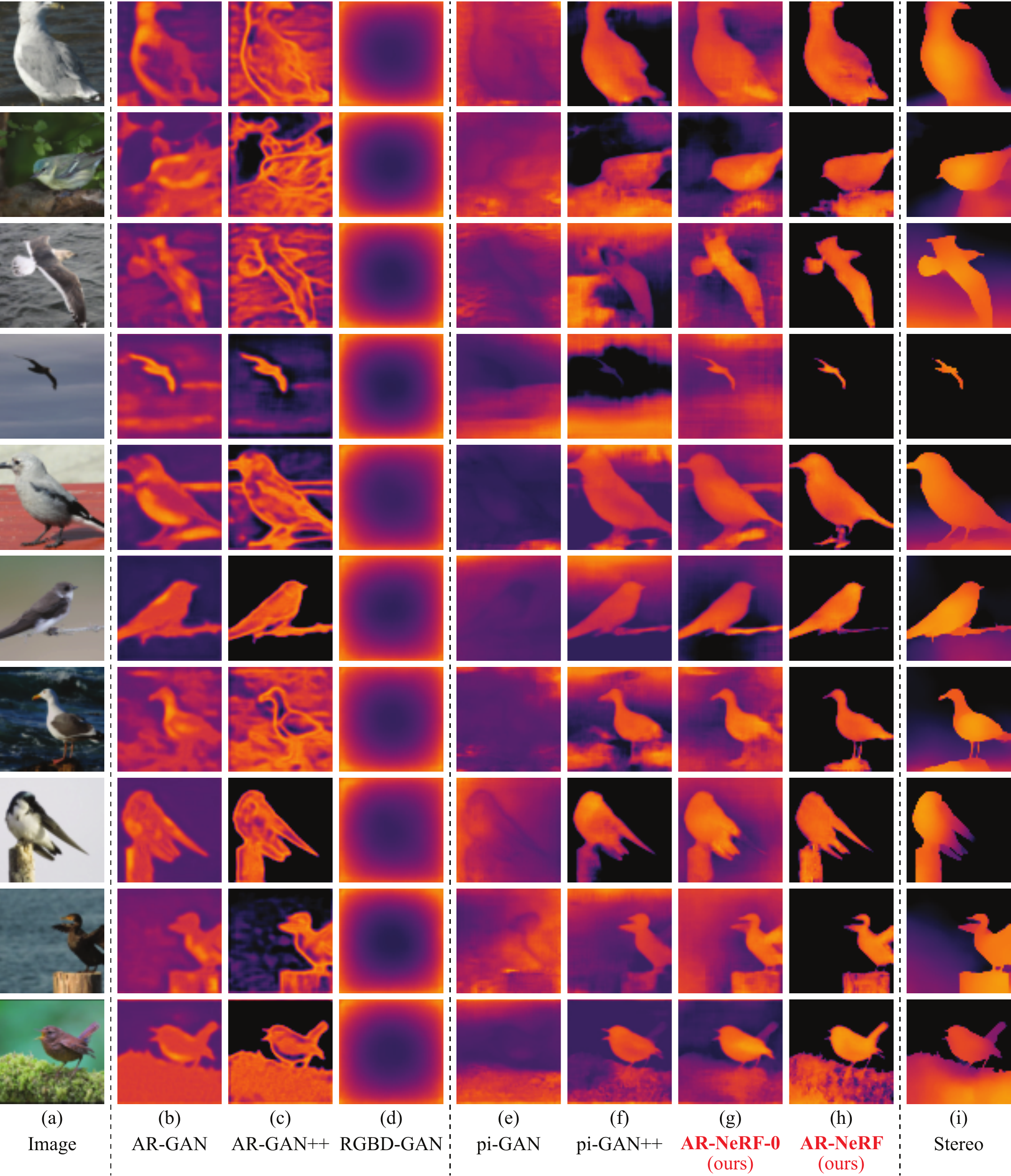}
  \caption{\textbf{Comparison of depth prediction on the CUB-200-2011 dataset.}
    These depths are used to calculate the SIDEs in Tables~\ref{tab:comparative_study} and \ref{tab:ablation_study}.
    AR-GAN (b), AR-GAN++ (c), and RGBD-GAN (d) are CNN-based and are trained in a \textit{fully} unsupervised manner.
    In addition, pi-GAN (e), pi-GAN++ (f), AR-NeRF-0 (g), and AR-NeRF (h) are NeRF-based and trained in a \textit{fully} unsupervised manner.
    By contrast, the model in (i)~\cite{KXianCVPR2020} was trained using stereo supervision and was applied as the \textit{ground truth} in the evaluation.
    The SIDEs in Tables~\ref{tab:comparative_study} and \ref{tab:ablation_study} were calculated by comparing the depths in (b)--(h) with the depths in (i).
    Our findings are summarized as follows:
    (1) Because AR-GAN (b) and AR-GAN++ (c) only employ defocus (appearance) cues, their predicted depths are affected by their appearance.
    For example, in the fifth row, the horizontal boundary in the background is emphasized despite its non-necessity.
    (2) RGBD-GAN (d) utilizes viewpoint (geometric) cues for 3D representation learning.
    However, there are few viewpoint cues in this dataset; consequently, this model has difficulty in learning depth.
    (3) For the same reason, pi-GAN (e) and pi-GAN++ (f), which also only employ viewpoint cues, suffer from learning difficulty, although NeRF itself has a strong 3D consistency at the design level.
    In particular, they fail to consistently distinguish between the foreground (bird) and background (surroundings), parts of which are often mixed (e.g., in the third and fourth rows).
    (4) Although the results of AR-NeRF-0 (g) are closest to those of AR-NeRF (h), AR-NeRF-0 (g) is often affected by the appearance (e.g., in the fifth row, similar to AR-GAN (b) and AR-GAN++ (c)) because it also only leverages focus cues.
    (5) AR-NeRF (h) overcomes the limitations of pi-GAN (e), pi-GAN++ (f), and AR-NeRF-0 (g) by utilizing both the viewpoint and defocus cues.}
  \label{fig:depth_prediction_birds}
\end{figure*}

\begin{figure*}[p]
  \centering
  \includegraphics[width=0.975\textwidth]{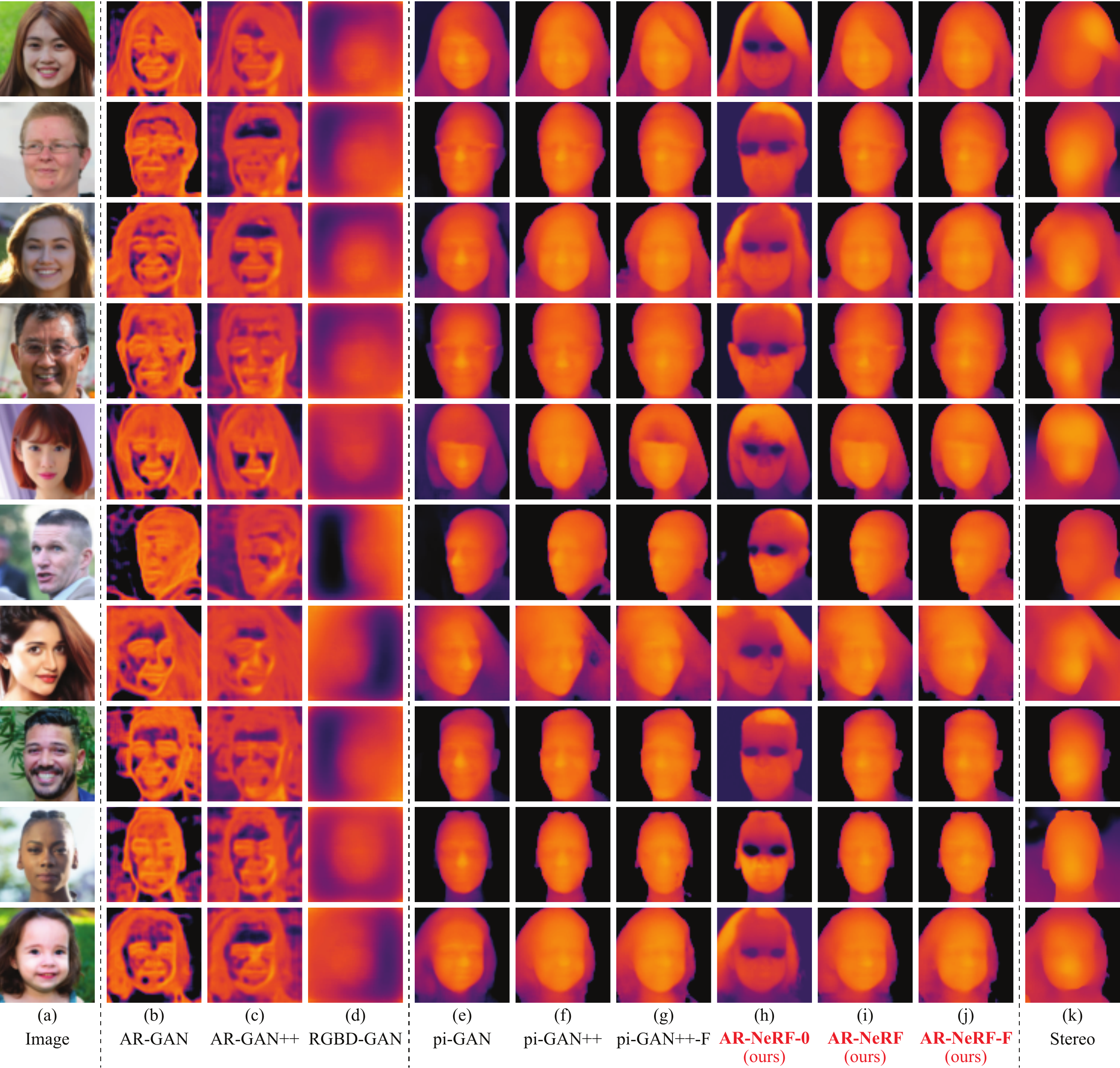}
  \caption{\textbf{Comparison of depth prediction on the FFHQ dataset.}
    These depths are used to calculate the SIDEs in Tables~\ref{tab:comparative_study} and \ref{tab:ablation_study}.
    AR-GAN (b), AR-GAN++ (c), and RGBD-GAN (d) are CNN-based and are trained in a \textit{fully} unsupervised manner.
    In addition, pi-GAN (e), pi-GAN++ (f), pi-GAN++-F (g), AR-NeRF-0 (h), AR-NeRF (i), and AR-NeRF-F (j) are NeRF-based and trained in a \textit{fully} unsupervised manner.
    By contrast, the model in (k)~\cite{KXianCVPR2020} was trained using stereo supervision and was applied as the \textit{ground truth} in the evaluation.
    The SIDEs in Tables~\ref{tab:comparative_study} and \ref{tab:ablation_study} were calculated by comparing the depths in (b)--(j) with those in (k).
    Our findings are summarized as follows:
    (1) In the depths predicted using AR-GAN (b) and AR-GAN++ (c), holes appeared in the face regions.
    This is because they can only adopt a defocus cue, which is insufficient to distinguish a flat surface in a face from the blur caused by the defocus.
    (2) RGBD-GAN succeeded in capturing the face direction in depth by leveraging the viewpoint cue (e.g., in the sixth and seventh rows); however, the depth fidelity was low.
    This can occur because RGBD-GAN imposes 3D consistency only at a loss level and not at an architectural level.
    (3) By contrast, the NeRF-based models (e)--(j) have 3D consistency at the architectural level, allowing high fidelity and consistent 3D depths to be predicted.
    In particular, we found that the models utilizing viewpoint cues ((e)--(g), (i), and (j)) demonstrated similar performance.
    This is because this dataset includes sufficiently varying viewpoints.
    (4) However, AR-NeRF-0 (h), which can only use the defocus cue, fails to capture structures around the eyes.
    This is possibly because there is a large variety of appearances around the eyes, and it is difficult to model the corresponding depth using only a defocus cue.
    We can overcome this limitation by jointly using viewpoint and defocus cues, as in AR-NeRF (i) or AR-NeRF-F (j).}
  \label{fig:depth_prediction_faces}
\end{figure*}

In this appendix, we provide additional qualitative results that correspond to those presented in the main text.
Figure captions and their relationship to the results in the main text are as follows:
\begin{itemize}
  \setlength{\parskip}{2pt}
  \setlength{\itemsep}{2pt}
\item Figure~\ref{fig:concept_extension}:
  Unsupervised learning of depth and defocus effects from unstructured (and view-limited) natural images.
  This figure is an extended version of Figure~\ref{fig:concept}.
\item Figure~\ref{fig:control_defocus_latent_flowers}:
  Simultaneous control of defocus and latent codes on the Oxford Flowers dataset.
  This figure is an extension of Figure~\ref{fig:concept}.
\item Figure~\ref{fig:control_defocus_latent_birds}:
  Simultaneous control of defocus and latent codes on the CUB-200-2011 dataset.
  This figure extends Figure~\ref{fig:concept}.
\item Figure~\ref{fig:control_defocus_latent_faces}:
  Simultaneous control of defocus and latent codes on the FFHQ dataset.
  This figure is an extended version of Figure~\ref{fig:concept}.
\item Figure~\ref{fig:generation_extension}:
  Comparison of generated images and depths between AR-GAN++ and AR-NeRF.
  This figure extends Figure~\ref{fig:generation}.
\item Figure~\ref{fig:depth_prediction_flowers}:
  Comparison of depth prediction on the Oxford Flowers dataset.
  The depths are used to calculate the SIDEs in Tables~\ref{tab:comparative_study} and \ref{tab:ablation_study}.
\item Figure~\ref{fig:depth_prediction_birds}:
  Comparison of depth prediction on the CUB-200-2011 dataset.
  The depths are used to calculate the SIDEs in Tables~\ref{tab:comparative_study} and \ref{tab:ablation_study}.
\item Figure~\ref{fig:depth_prediction_faces}:
  Comparison of depth prediction on the FFHQ dataset.
  The depths are used to calculate the SIDEs in Tables~\ref{tab:comparative_study} and \ref{tab:ablation_study}.
\end{itemize}

\clearpage
\section{Implementation details}
\label{sec:implementation_details}

In this appendix, we provide implementation details regarding the following items:

\begin{itemize}
  \setlength{\parskip}{2pt}
  \setlength{\itemsep}{2pt}
\item Appendix~\ref{subsec:implementation_main}:
  Details of the main experiments (Section~\ref{sec:experiments}).
\item Appendix~\ref{subsec:implementation_defocus_renderer}:
  Details of defocus renderer (Appendix~\ref{subsec:application}).
\end{itemize}

\subsection{Details of main experiments (Section~\ref{sec:experiments})}
\label{subsec:implementation_main}

\subsubsection{Dataset}

In the experiments, we used three datasets,
the detailed information of which is as follows:

\smallskip\noindent\textbf{Oxford Flowers~\cite{MENilsbackICVGIP2008}.}
The dataset consists of 8,189 images with 102 flower categories.
Each category includes 40 or more images.
The images were obtained by searching the web and taking photographs.
We downloaded the data from an official website.\footnote{\url{https://www.robots.ox.ac.uk/~vgg/data/flowers/102/}.}
More detailed information is provided in the README file available on the website.

\smallskip\noindent\textbf{CUB-200-2011~\cite{CWahCUB2002011}.}
The dataset contains 11,788 images of 200 bird species.
The images were collected using a Flickr image search and then filtered by presenting each image to multiple users of Mechanical Turk~\cite{PWelinderNIPS2010}.
We downloaded the data from an official website.\footnote{\url{http://www.vision.caltech.edu/visipedia/CUB-200-2011.html}.}
More detailed information is provided in the technical report~\cite{CWahCUB2002011}.

\smallskip\noindent\textbf{FFHQ (Flickr-Faces-HQ)~\cite{TKarrasCVPR2019}.}
The dataset consists of 70,000 face images.
The images were crawled from Flickr.
Therefore, as the dataset creators~\cite{TKarrasCVPR2019} mention, the dataset inherits all the biases of that website.
The images were filtered using automatic filters and Amazon Mechanical Turk.
Only images under permissive licenses (Creative Commons BY 2.0, Creative Commons BY-NC 2.0, Public Domain Mark 1.0, Public Domain CC0 1.0, or U.S. Government Works license) were collected.
The dataset itself is available under a Creative Commons BY-NC-SA 4.0 license by the NVIDIA Corporation.
We downloaded the data from an official website.\footnote{\url{https://github.com/NVlabs/ffhq-dataset}.}
More detailed information is provided in the README file available on the website.

\subsubsection{Network architectures}

As explained in Section~\ref{subsec:experimental_settings}, we implemented AR-NeRF based on the pi-GAN~\cite{EChanCVPR2021},\footnote{\url{https://github.com/marcoamonteiro/pi-GAN}.} which is a state-of-the-art generative variant of NeRF.
Because the original pi-GAN was not applied to the datasets used in our experiments, we carefully tuned the configurations and hyperparameters so that the baseline pi-GAN could generate images sufficiently well.
In particular, we used the configuration of CelebA~\cite{ZLiuICCV2015}\footnote{\label{foot:celeba_config}\url{https://github.com/marcoamonteiro/pi-GAN/blob/master/curriculums.py}.} as the default and tuned depending on the dataset.
We explain the details of each network below.

\smallskip\noindent\textbf{Mapping network.}
In pi-GAN, a StyleGAN~\cite{TKarrasCVPR2019}-inspired mapping network was introduced to efficiently propagate information in the latent code to each layer.
We implemented this network using an MLP with three hidden layers (256 units each).
We used leaky rectified linear units (LReLUs)~\cite{AMaasICML2013} with a negative slope of 0.2 as activation functions.
The dimension of the latent code was set to 256.
This architecture is the same as that of the original pi-GAN~\cite{EChanCVPR2021}.

\smallskip\noindent\textbf{Synthesis network.}
In pi-GAN, SIREN~\cite{VSitzmannNeurIPS2020}-based implicit radiance fields are used as a synthesis network.
We implemented this network using an MLP with eight FiLM~\cite{EPerezAAAI2018,VDumoulinDistill2018}-SIREN hidden layers of 128 units each.
In the original pi-GAN~\cite{EChanCVPR2021}, 256 units were used in each layer; however, in our preliminary experiments, we found that the reduction in the units did not significantly affect the performance.
In addition, this reduction allowed the use of a larger number of points along the ray, which is critical for improving performance.
Considering this, we used 128 units in our experiments.
In pi-GAN++ and AR-NeRF, we used the above-mentioned network as a foreground synthesis network and implemented a background synthesis network using an MLP with eight FiLM-SIREN hidden layers of 64 units each.
We used fewer parameters in the background synthesis network under the assumption that the background is simpler than the foreground.

\smallskip\noindent\textbf{Discriminator.}
We used different discriminators according to the dataset.
For FFHQ, we used the same discriminator as that used in pi-GAN for CelebA.\footnoteref{foot:celeba_config}
The discriminator was implemented using CoordConv layers~\cite{RLiuNeurIPS2018} and residual blocks~\cite{KHeCVPR2016}.
In our preliminary experiments, we found that the CoordConv layers yield negative effects for Oxford Flowers and CUB-200-2011.
A possible cause is that in FFHQ, faces are aligned based on facial landmarks, whereas in Oxford Flowers and CUB-200-2011, flowers and birds are not strictly aligned.
Based on this finding, we removed the CoordConv layers from the discriminator when applied to Oxford Flowers and CUB-200-2011.

\subsubsection{Training settings}

We used different training settings according to the dataset.
For FFHQ, we used the same setting as that in pi-GAN for CelebA.\footnoteref{foot:celeba_config}
More specifically, as the GAN objective function, we used the non-saturating GAN loss~\cite{IGoodfellowNIPS2014} with real gradient penalty ($R_1$) regularization~\cite{LMeschederICML2018}, where the weight parameter of the $R_1$ regularization was set to 0.2.
Additionally, we used an identity regularizer~\cite{TNguyenICCV2019} with the weight parameter of 15 to keep the identity across different viewpoints.
The network was trained for 200,000 iterations using the Adam optimizer~\cite{DPKingmaICLR2015}, with learning rates of 0.00006 and 0.0002 for the generator and discriminator, respectively, and momentum terms $\beta_1$ and $\beta_2$ of 0 and 0.9, respectively.
The batch size was set to 16.
We used an exponential moving average~\cite{TKarrasICLR2017} with a decay of 0.999 over the weights to generate the final generator.
In pi-GAN, we set the number of sample points along the ray to 48, where 32 and 16 points were used for stratified sampling and hierarchical sampling, respectively.
In pi-GAN++ and AR-NeRF, we set the values for the foreground and background synthesis networks as 48 and 24, respectively.
In the foreground synthesis, 32 and 16 points were used for stratified sampling and hierarchical sampling, respectively.
In the background synthesis, 16 and 8 points were used for stratified sampling and hierarchical sampling, respectively.
A field of view was set to $12^{\circ}$.
As discussed in Section~\ref{subsec:ablation_study}, we used the same number of rays in all models to investigate the pure performance differences between the models with and without aperture rendering.
Specifically, we used five stratified sampled rays (Section~\ref{subsec:advanced_techniques}) in AR-NeRF and five ensemble rays (i.e., five rays with an aperture size $s = 0$) in pi-GAN and pi-GAN++.

For Oxford Flowers and CUB-200-2011, we also used differentiable augmentation~\cite{SZhaoNeurIPS2020}\footnote{\url{https://github.com/mit-han-lab/data-efficient-gans}.} to stabilize the training.
In particular, we used color jittering, translation, and cutout ~\cite{TDevriesArXiv2017} for Oxford Flowers, and translation for CUB-200-2011 because we found that they were the best choice.
We removed the identity regularizer~\cite{TNguyenICCV2019} because we found that it yields negative effects for Oxford Flowers and CUB-200-2011.
The other settings were the same as those for FFHQ.

\subsubsection{Evaluation}

As described in Section~\ref{subsec:experimental_settings}, we calculated KID using 20,000 generated images and all real images.
We implemented a KID calculator based on the official code.\footnote{\url{https://github.com/mbinkowski/MMD-GAN}.}
The depth predictor, which was used when calculating the SIDE, was implemented using U-Net architecture~\cite{ORonnebergerMICCAI2015}.
In particular, we used the same network and training settings as those used in the AR-GAN study~\cite{TKanekoCVPR2021b} for direct comparison.
pi-GAN++ and AR-NeRF can synthesize the unbounded background (or depth) by using a NeRF++~\cite{KZhangArXiv2020}-based background synthesis network; however, the predictable depth range is bounded in a typical depth predictor, including the model~\cite{KXianCVPR2020} that was used as ``ground truth'' in our experiment.
Therefore, we used only the foreground synthesis network when calculating the SIDE.

\subsection{Details of defocus renderer (Appendix~\ref{subsec:application})}
\label{subsec:implementation_defocus_renderer}

\subsubsection{Dataset}
In the experiment, we used a test set of the \textit{iPhone2DSLR Flower}~\cite{JYZhuICCV2017} for the evaluation.
Its detailed information is as follows:

\smallskip\noindent\textbf{iPhone2DSLR Flower~\cite{JYZhuICCV2017}.}
The dataset includes 2,381 smartphone images and 3,805 DSLR images.
The smartphone images were collected from Flickr by searching for photos taken by Apple iPhone 5, 5s, or 6, with the search text ``flower.''
DSLR images with a shallow depth-of-field (DoF) were also collected from Flickr using the search tags ``flower'' and ``dof.'' 
We downloaded the data from an official website.\footnote{\url{https://github.com/junyanz/CycleGAN}.}
More detailed information is provided on the website and appendix of the corresponding paper~\cite{JYZhuICCV2017}.

\subsubsection{Network architectures}

We implemented AR-NeRF-R using basically the same network as AR-GAN-DR~\cite{TKanekoCVPR2021b}, that is, we used the U-Net architecture~\cite{ORonnebergerMICCAI2015}.
A difference from AR-GAN-DR is that we extended U-Net to a conditional setting~\cite{JYZhuNIPS2017}.
Specifically, we injected the aperture size $s$ and focus distance $f$ into every intermediate layer in the encoder after expanding them to the corresponding feature map size.

\subsubsection{Training settings}

We generated training data (i.e., pairs of all-in-focus and focused images with auxiliary information on aperture size $s$ and focus distance $f$) using AR-NeRF, which was trained using $64 \times 64$ images on the Oxford Flowers dataset.
AR-NeRF was the same as that used to generate the samples in Figures~\ref{fig:concept} and \ref{fig:generation}.
When training AR-NeRF-R, we used $128 \times 128$ images generated by AR-NeRF, where we increased the resolution of the generated images from $64 \times 64$ to $128 \times 128$ by increasing the density of input points (Section~\ref{subsec:comparative_study}).
We trained the defocus renderer for 300,000 iterations using the Adam optimizer~\cite{DPKingmaICLR2015} with a learning rate of 0.0003 and momentum terms $\beta_1$ and $\beta_2$ of 0.9 and 0.99, respectively.
The batch size was set to 4.
The learning rate was kept constant during training, except for the last $30\%$ iterations, where the learning rate was smoothly ramped down to zero.

\end{document}